# Surrogate modeling for interpreting black-box LLMs in medical predictions

Changho Han, MD PhD[1¶], Songsoo Kim, MD[2¶], Dong Won Kim, MS[2¶], Leo Anthony Celi, MD MS MPH[3,4,5], Jaewoong Kim, MS[2], SungA Bae, MD PhD[6,7]*, Dukyong Yoon, MD PhD[2,7,8]*

[1]Medical Big Data Research Center, Seoul National University Medical Research Center, Seoul National University College of Medicine, Seoul, Republic of Korea

[2]Department of Biomedical Systems Informatics, Yonsei University College of Medicine, Seoul, Republic of Korea.

[3]Laboratory for Computational Physiology, Massachusetts Institute of Technology, Cambridge, MA 02139

[4]Division of Pulmonary, Critical Care and Sleep Medicine, Beth Israel Deaconess Medical Center, Boston, MA 02215

[5]Department of Biostatistics, Harvard T.H. Chan School of Public Health, Boston, MA 02115

[6]Department of Cardiology, Yongin Severance Hospital, Yonsei University College of Medicine, Yongin, Republic of Korea.

[7]Center for Digital Health, Yongin Severance Hospital, Yonsei University Health System, Yongin, Republic of Korea.

[8]Institute for Innovation in Digital Healthcare, Severance Hospital, Seoul, Republic of Korea.

¶These authors contributed equally to this study and share first authorship.

*Corresponding authors:

Dukyong Yoon: dukyong.yoon@yonsei.ac.kr; 331 Dongbaekjukjeon-daero, Giheung-gu, Yongin-si, Gyeonggi-do 16995, Republic of Korea. Tel.: 82-31-5189-8450

SungA Bae: cardiobsa@yuhs.ac; 363 Dongbaekjukjeon-daero, Giheung-gu, Yongin-si, Gyeonggi-do 16995, Republic of Korea. Tel.: 82-31-5189-8958

# Abstract

Large language models (LLMs), trained on vast datasets, encode extensive real-world knowledge within their parameters, yet their black-box nature obscures the mechanisms and extent of this encoding. Surrogate modeling, which uses simplified models to approximate complex systems, can offer a path toward better interpretability of black-box models. We propose a surrogate modeling framework that quantitatively explains LLM-encoded knowledge. For a specific hypothesis derived from domain knowledge, this framework approximates the latent LLM knowledge space using observable elements (input-output pairs) through extensive prompting across a comprehensive range of simulated scenarios. Through proof-of-concept experiments in medical predictions, we demonstrate our framework's effectiveness in revealing the extent to which LLMs "perceive" each input variable in relation to the output. Particularly, given concerns that LLMs may perpetuate inaccuracies and societal biases embedded in their training data, our experiments using this framework quantitatively revealed both associations that contradict established medical knowledge and the persistence of scientifically refuted racial assumptions within LLM-encoded knowledge. By disclosing these issues, our framework can act as a red-flag indicator to support the safe and reliable application of these models.

# INTRODUCTION

Large language models (LLMs), trained on vast datasets, encode extensive real-world knowledge within their parameters[1,2]. However, as black-box models, this knowledge is implicitly distributed across the network in an inaccessible format, limiting explainability and hindering understanding of encoding mechanisms, specific knowledge extent, and inter-knowledge relationships[3-5]. This poses a significant barrier in many domains, where understanding the basis of model outputs is essential for fostering trust and ensuring accountability[3,5,6]. Additionally, as LLMs are increasingly utilized, challenges like hallucinations and bias have become critical[7-12]. For instance, Zack et al.[7] found that GPT-4's treatment recommendations, such as referrals and imaging, varied by race and gender even when clinical presentations were identical, while Abid et al.[8] demonstrated that GPT-3 disproportionately associates Muslims with violence-related terms. However, the black-box nature of LLMs hinders the ability to fully understand and address the extent of these issues. These challenges are especially critical in high-stakes domains such as healthcare, where biased or inaccurate outputs can impact patient outcomes and exacerbate disparities.

Surrogate modeling, which employs simplified representations to approximate complex, high-order systems, has been applied to black-box models to achieve better interpretability and computational efficiency[13,14]. By exploring the application of surrogate modeling to foundation LLMs, which exemplify one of the most complex cutting-edge artificial intelligence (AI) systems, we may gain better insights into their specific knowledge, biases, and inaccuracies.

In this context, we propose a surrogate modeling framework to quantitatively explain LLM-encoded knowledge for specific hypotheses. As a proof-of-concept, we present results from experiments conducted in the context of medical predictions, demonstrating how surrogate modeling can elucidate LLM-encoded knowledge related to a target hypothesis, thereby quantifying the extent of potential biases and inaccuracies.

# RESULTS

## Overview of the surrogate modeling framework

The Methods section provides a detailed breakdown of our framework's surrogate modeling approach; here, we briefly outline the overall concept and present an overview of the framework's design.

Our surrogate modeling framework can be summarized as the quantification of LLM input-output pairs for a target hypothesis derived from domain knowledge, achieved through extensive prompting across a wide range of simulated scenarios (Fig. 1). It consists of three stages (Fig. 2). The first stage involves generating a comprehensive and extensive simulated dataset for the target hypothesis, following these steps: outcome selection, input variable selection, probability distribution selection, and random sampling. The choice of input variables (and their probability distributions) related to the selected outcome is guided by the target hypothesis derived from prior domain knowledge. The second stage involves extensively obtaining LLM responses for the broad and diverse spectrum of simulated scenarios generated in the first stage. The third stage involves applying statistical modeling (e.g., linear regression), also guided by the target hypothesis, to the pairs of simulated data and LLM responses to quantify and generalize LLM-encoded knowledge into a single statistical formula, making it explainable. In our experiments, we chose simple statistical models (i.e., linear regression and log-transformed linear regression) for their simplicity and intuitive interpretability. A detailed methodological breakdown of our framework is provided in the Methods section.

The following sections will detail the four proof-of-concept experiments we conducted to validate our framework in the context of medical predictions.

## Experiment 1: Cardiovascular disease risk prediction

A previous study showed that LLMs perform comparably to traditional risk prediction tools in cardiovascular disease (CVD) risk prediction[15]. Therefore, the first experiment applied surrogate modeling to LLMs for CVD risk prediction, aiming to demonstrate the framework's effectiveness in explaining LLM-encoded knowledge.

### Stage 1. Simulated dataset generation

*Stage 1-1. Outcome selection*

We selected CVD risk prediction as the topic for our first experiment. CVD poses a major health burden[16], and various treatment guidelines recommend CVD risk prediction tools such as the American College of Cardiology/American Heart Association Pooled Cohort Equation (PCE)[17] to guide CVD risk management, including decisions on blood pressure and statin treatments based on 10-year CVD risk estimates[18].

*Stage 1-2. Input variable selection*

We searched for well-established CVD risk factors through an extensive literature review (Supplementary Method S1). As a result, a total of 20 well-established CVD risk factors were selected (Supplementary Table S1): age, sex, diabetes, anti-hypertensive treatment, dyslipidemia, atrial fibrillation, chronic kidney disease (CKD), family history of CVD in first degree relatives (FHCVD), smoking status, systolic blood pressure, diastolic blood pressure, body mass index, waist-hip ratio, high-density lipoprotein cholesterol, low-density lipoprotein cholesterol, triglycerides, hemoglobin A1c (HbA1c), creatinine, uric acid, and C-reactive protein. To test the flexibility of our framework in various scenarios, we also created and tested multiple input variable combinations.

*Stage 1-3. Probability distribution selection*

We set a probability distribution for each input variable. For continuous variables, we set the probability distribution as a lognormal distribution, as many studies have demonstrated that this approach is physiologically appropriate[19-21]. We set the lognormal distribution broadly enough to reflect both physiologically normal and abnormal states (Supplementary Method S2 and Supplementary Fig. S1). For categorical variables, assuming a sufficiently large dataset, the proportion of samples in each category should not influence the regression coefficients, so we allocated equal distribution across categories.

*Stage 1-4. Random sampling*

Subsequently, we performed random sampling of 20,000 cases from the predefined distribution set for each variable to generate a simulated dataset.

**Stage 2. Obtaining LLM responses**

Next, we structured each sample of the simulated dataset into sentence form to create LLM prompts to elicit 10-year CVD risk estimations (Supplementary Table S2). Three foundation LLMs were used in a zero-shot setting: GPT-4[22], Claude-3.5-Sonnet[23], and Gemini-1.5-Pro[24]. For each sample, we collected 10 responses from each of the three LLMs.

**Stage 3. Statistical modeling**

Finally, we used the 20,000*10 pairs of input variables and LLM-predictions to establish a linear regression model, where the LLM-prediction served as the dependent variable and the input variables acted as independent variables, culminating in the surrogate prediction model.

**Interpretation of the surrogate model**

The encoded knowledge of LLMs regarding each variable's association with CVD risk prediction was quantified (Fig. 3, Supplementary Table S1). The directionality (sign) of most coefficients was consistent across the three LLM models; however, differences were observed in the magnitude of the coefficients, and for some coefficients, even the directionality (sign) was inconsistent among the models. For example, the coefficient for waist-hip ratio was 9.3 times larger in Claude-3.5-Sonnet compared to GPT-4 (Waist-hip ratio coefficient: GPT-4, 0.426 [95% confidence interval, CI: 0.137–0.714]; Claude-3.5-Sonnet, 3.948 [95% CI: 3.734–4.162]). In contrast, the coefficient for waist-hip ratio in Gemini-1.5-Pro was statistically insignificantly negative (-0.313 [95% CI: -0.660 – 0.034]). Similarly, the directionality of the coefficients for uric acid differed across models, being positive in GPT-4 and Claude-3.5-Sonnet but negative in Gemini-1.5-Pro (Uric acid coefficient: GPT-4, 0.035 [95% confidence interval, CI: 0.014–0.056]; Claude-3.5-Sonnet, 0.053 [95% CI: 0.038–0.069]; Gemini-1.5-Pro, -0.046 [95% CI: -0.071 – -0.021]). Supplementary Table S3 shows the linear regression coefficients for various input variable combinations for GPT-4, with similar trends observed across all combinations.

**Evaluation of the surrogate model**

Using data from the United Kingdom Biobank (UKB)[25], we evaluated the predictive ability of the surrogate prediction model for major adverse cardiovascular events (MACEs). Among 357,113 individuals analyzed from the UKB (Supplementary Fig. S2), 22,866 (6.4%) experienced MACEs within 10 years (Supplementary Table S4). Differences were observed in the distribution of predicted risks across the PCE, FRS, and surrogate models for GPT-4, Claude-3.5-Sonnet, and Gemini-1.5-Pro (Supplementary Table S5). The FRS and the surrogate model for GPT-4 produced relatively high mean risk estimates (14.2% and 13.7%, respectively), whereas the PCE and the Gemini-1.5-Pro surrogate model yielded lower mean estimates (both 7.9%). The surrogate model for Claude-3.5-Sonnet showed intermediate predictions, with a mean of 11.8%. The surrogate model for GPT-4, which achieved a Harrell's concordance index (C-index) of 0.736, surpassed the PCE by 0.010 (95% CI: 0.008–0.011) and the Framingham

Risk Score (FRS) by 0.016 (95% CI: 0.014–0.018) (Fig. 4, Supplementary Table S6)[17,26]. In experiments using various input variable combinations, the C-index for the surrogate model improved as more relevant CVD risk factors were included (Supplementary Fig. S3). Similar trends were observed in experiments using Claude-3.5-Sonnet and Gemini-1.5-Pro; the surrogate model for Claude-3.5-Sonnet, with a C-index of 0.740, outperformed the PCE by 0.014 (95% CI: 0.012–0.015) and the FRS by 0.020 (95% CI: 0.018–0.022); the surrogate model for Gemini-1.5-Pro, with a C-index of 0.732, outperformed the PCE by 0.005 (95% CI: 0.003–0.007) and the FRS by 0.011 (95% CI: 0.010–0.013).

By applying the SHapley Additive exPlanations (SHAP) method to the surrogate prediction model[27], the contribution of each variable to the output was explained at the individual level—an insight that cannot be obtained directly from the LLM itself (Fig. 5). Moreover, the surrogate model yielded deterministic outputs for a given sample, whereas directly querying LLMs resulted in inconsistent responses. In addition, inference using the surrogate model was significantly faster—approximately 900 to 2100 times—than obtaining responses from each LLM via application programming interface (API)-based prompting (Supplementary Table S7). As a demonstration of making the surrogate model fully accessible, we created an online tool at https://llm-surrogate.com/[28].

## Experiment 2: Estimation of glomerular filtration rate

A previous study has shown that when LLMs were directly asked how glomerular filtration rate (GFR) is estimated, they often generated responses based on scientifically refuted race-based equations[29]. Thus, the second experiment applied surrogate modeling to LLMs for GFR estimation, aiming to further validate the framework by quantifying potential biases in LLM-encoded knowledge.

GFR is a measure of how well the kidneys are filtering blood, which can be estimated using variables such as age, sex, and serum creatinine (Scr) through the Chronic Kidney Disease Epidemiology Collaboration (CKD-EPI) Equation (estimated GFR, eGFR)[30]. Historically, this equation included a race-based adjustment[31], but scientific consensus has refuted its use, recognizing that race is a social construct rather than a biological one[30,32].

Therefore, in “Stage 1-2: Input variable selection”, we included age, sex, Scr, and race (White or Black) as input variables to quantitatively investigate whether the scientifically refuted racial bias is still present in the LLM-encoded knowledge. We followed the same principles as in the previous experiment to generate a simulated dataset of 20,000 samples, prompting the LLMs to provide GFR estimations. For each sample, we collected a single response from each of the three LLMs. Subsequently, we conducted statistical modeling using the same formula as in the CKD-EPI equation (equation (1))[30,31].

CKD-EPI formula:

$$eGFR = \mu * \min\left(\frac{Scr}{\alpha}, 1\right)^{\beta_1} * \max\left(\frac{Scr}{\alpha}, 1\right)^{\beta_2} * \gamma^{Age} * \delta[if\ Female] * \varepsilon[if\ Black] \quad (1)$$

Linear regression:

$$\log eGFR = \log\mu + \beta_1 * \log\left(\min\left(\frac{Scr}{\alpha}, 1\right)\right) + \beta_2 * \log\left(\max\left(\frac{Scr}{\alpha}, 1\right)\right) + \log\gamma * Age + \log\delta * Female + \log\varepsilon * Black \quad (2)$$

We applied a log transformation to both sides of the CKD-EPI formula and conducted linear regression (equation (2), $\alpha$ is 0.7 for females and 0.9 for males, “min” indicates the minimum of the two values and “max” indicates the maximum). To align with the CKD-EPI equations[32,33], we calculated $\beta_1$ separately for each sex.

The trends for age, creatinine and sex in GFR estimation were similar across the three LLMs, though the coefficients varied, reflecting the differences in the knowledge each model possesses (Fig. 6, Supplementary Table S8). There were also differences in the coefficients for Black race among the models. In the experiment with GPT-4, the coefficient for Black race was 1.153 (95% CI: 1.149–1.158), indicating GPT-4 estimates GFR values to be 15.3% higher for Black individuals compared to White individuals. This bias was even more pronounced in Claude-3.5-Sonnet, where the coefficient was 1.227 (95% CI: 1.224–1.230), exceeding the value of 1.159 (95% CI: 1.144–1.170) in the refuted, race-based 2009 CKD-EPI eGFRcr equation[30,31], suggesting potentially stronger racial bias in this model. In contrast, Gemini-1.5-Pro showed a much smaller coefficient for Black race at 1.008 (95% CI: 1.005–1.010).

Differences were observed in the distribution of estimated GFR across the CKD-EPI eGFRcr formulas, the direct estimations from each LLM, and the corresponding surrogate models within the simulated dataset (Supplementary Table S9). GPT-4 and Claude-3.5-Sonnet showed a tendency to overestimate GFR compared to the CKD-EPI formulas. Specifically, the mean GFR values from GPT-4 and its surrogate model were 111.7 and 111.0 mL/min/1.73m$^2$, respectively, while those from Claude-3.5-Sonnet and its surrogate model were 108.9 and 108.8 mL/min/1.73m$^2$. In contrast, the 2021 and 2009 CKD-EPI eGFRcr formulas produced lower mean estimates of 92.7 and 97.4 mL/min/1.73m$^2$, respectively. Estimations from Gemini-1.5-Pro and its surrogate model (96.2 and 96.0 mL/min/1.73m$^2$) fell between the two CKD-EPI estimates. As shown in Supplementary Fig. S4, the direct estimations from each LLM exhibited a tendency to cluster around specific values. In contrast, the estimations from the corresponding surrogate models demonstrated smoother, more continuous distributions.

### Experiments 3 & 4

To further demonstrate the generalizability of the surrogate modeling framework across different subjects, we conducted two additional experiments. The third experiment focused on stroke risk prediction (Supplementary Result S1). Surrogate modeling effectively quantified the association of each input variable with the outcome. The directionality of the coefficients was consistent across the three LLM models, though differences were observed in their magnitudes (Supplementary Fig. S5 and Supplementary Table S10). Additionally, the constructed surrogate model exhibited performance comparable to a traditional risk prediction tool, while the distributions of predicted risks showed differences (Supplementary Result S1, Supplementary Table S11). The fourth experiment, focusing on estimating pulmonary function (Supplementary Result S2), quantitatively confirmed the persistence of scientifically refuted racial biases in LLM-encoded knowledge (Supplementary Fig. S6 and Supplementary Table S12). Differences were observed in the distributions of estimated pulmonary function values across direct LLM estimations, the corresponding surrogate models, and a reference equation within the simulated dataset (Supplementary Table S13). Direct estimations from each LLM tended to cluster at specific values, while those from the surrogate models exhibited smoother distributions (Supplementary Fig. S7). Detailed information on these two experiments is included in Supplementary Results S1 and S2.

## DISCUSSION

We introduced a surrogate modeling framework for explaining LLM-encoded knowledge. It quantifies LLM input-output pairs for a target hypothesis derived from domain knowledge by extensively prompting the model using a wide-ranging simulated dataset. Proof-of-concept experiments in medical predictions demonstrated the framework's effectiveness in elucidating LLM-encoded knowledge and quantifying potential biases and inaccuracies.

Recent AI research has prioritized performance, often overlooking interpretability and understanding[3,5,6]. Unlike classical machine/statistical learning, which starts with hypotheses based on domain knowledge and learns from data, the availability of powerful computational resources, vast datasets, and advanced optimization algorithms has shifted the focus toward performance-oriented, empirically-driven search strategies in modern AI algorithms[3]. LLMs exemplify such overparameterized systems, with state-of-the-art models containing billions or trillions of parameters. In these models, the direct link between the hypothesis (knowledge) space and its representation is effectively lost (Fig. 1)[3], making it practically impossible to identify knowledge encoded by individual parameters. Instead, the LLM-encoded knowledge should be understood as an aggregate of parameter-wise representations, observable indirectly through responses to prompts. Given this, we hypothesized and confirmed that while the properties of the LLM knowledge space cannot be explicitly known[3,4], they can instead be approximated via surrogate modeling using observable elements (input-output pairs) for specific hypotheses derived from domain knowledge. These hypotheses guide the elicitation of LLM input-output pairs and the statistical modeling of these pairs. This statistical modeling aligns observed input-output pairs with domain-based hypotheses, quantifying relationships and identifying patterns in the LLM's knowledge space. The statistical formula reflects the degree to which these models "perceive" each input variable in relation to the desired outcome.

Our surrogate modeling framework enables quantitative assessment of LLM-encoded knowledge, helping identify potential inaccuracies or biases, and ensure safe, reliable applications. While a higher waist-hip ratio is an established independent risk factor for CVD[33,34], our experiments revealed variations in how each LLM "perceives" the relationship between waist-hip ratio and CVD risk prediction. Notably, one LLM even demonstrated a statistically insignificant negative association, indicating that it may have encoded incorrect or inconsistent knowledge. Similarly, while the association between elevated uric acid levels and CVD is well-known[35-37], our experiments uncovered an inverse relationship in the encoded knowledge of one LLM.

Furthermore, concerns have been raised that LLMs may encode and perpetuate societal biases from training data, compromising fairness and inclusivity[7-11]. A previous study[29] has shown that when LLMs were directly asked how GFR and lung capacity are estimated, they often generated responses based on scientifically refuted race-based equations[29,30,32,38]. Specifically, Omiye et al.[29] found that multiple LLMs continued to recommend race-adjusted eGFR calculations even though such racial assumptions had been formally refuted and their use officially discontinued by medical societies[30,32]. Similarly, these models suggested race-based spirometry adjustments despite professional guidelines explicitly rejecting such practices[38]. While these findings suggest that LLMs may contain inherent biases, relying solely on a few responses inadequately reveals the full extent of these biases. We have demonstrated that the surrogate modeling approach addresses this gap by enabling the systematic quantification of such biases into statistically interpretable values, offering a clearer view of how they manifest in LLM-encoded knowledge. Through this quantitative assessment, we found that while some LLMs exhibit minimal racial bias for GFR estimation, others display an even greater degree of bias than the refuted race-based equations. Overestimation of eGFR in Black individuals can obscure kidney impairment severity, delaying CKD diagnosis and treatment, exacerbating health disparities[30]. Where biases are detected, mitigation strategies—such as targeted reinforcement learning with human feedback[39,40]—may be needed for safe deployment.

While our study's main goal was to explain LLM-encoded knowledge through surrogate modeling, establishing the surrogate model's performance is essential to ensure the relevance of its interpretability. Poor predictive performance would render insights from the surrogate model irrelevant, as they would reflect a failed approximation. Our CVD risk prediction experiments showed that the surrogate model outperformed traditional CVD risk prediction tools on real-world data, confirming its relevance in both interpretability and predictive accuracy. One thing to note is that if LLMs merely replicated validated clinical formulas, the added value of both the LLMs and their corresponding surrogate models would be limited; however, across all our experiments, we observed clear differences in the distributions of predictions among LLMs, surrogate models, and established reference equations.

The surrogate modeling approach for LLMs offers several advantages over directly using the LLM. Previous studies have identified significant limitations when directly querying LLMs, including response inconsistencies across multiple runs[41-43]. Han et al.[15] discovered a distinct phenomenon termed "streaking", where CVD risk predictions via GPT-4 collapse into narrow diagonal clusters rather than producing a continuous distribution. This undesirable clustering behavior, combined with the run-to-run inconsistencies documented across medical domains, underscores the need for more robust approaches to extracting knowledge from LLMs. These challenges are effectively addressed through a generalized statistical formula of LLM-encoded knowledge. In our study as well, we observed that direct estimations from LLMs tended to cluster around specific values. In contrast, the corresponding surrogate model predictions exhibited smoother and more continuous distributions, further supporting the ability of surrogate modeling to mitigate "streaking" effects and enhance reliability. Moreover, while understanding the basis of model predictions is crucial for building trust and facilitating informed decision-making, unlike directly querying LLMs, our surrogate modeling approach explains, for every result generated, the extent to which LLMs "perceive" the contribution of each variable to the output. In contrast to recent prompt-based explainability techniques for medical LLMs—such as hierarchical prompting methods[44] and clinical reasoning frameworks[45]—which are highly sensitive to prompt variations, with even minor changes in wording significantly altering the results, our surrogate modeling approach requires minimal reliance on prompt formulations and, once derived through knowledge distillation, provides explanations in a lightweight and deterministic way without any further LLM inference requirements. Moreover, compared to direct LLM inference, surrogate models significantly reduce both inference time and computational cost, thereby improving cost-benefit profiles even in resource-constrained hospital environments[46]. As a demonstration of making the surrogate model fully accessible, we created an online tool at https://llm-surrogate.com/[28].

Users can define target hypotheses from domain knowledge for LLM surrogate modeling and specify inputs to generate simulated datasets accordingly. Our surrogate modeling framework accommodates variation in elements such as variable selection, the probability distributions used to generate the simulated dataset, the number of total prompts, or the number of responses collected per prompt. This user-driven approach to simulated dataset creation offers flexibility and easy re-implementation, in contrast to traditional prediction models that rely on a fixed set of risk factors and face update challenges requiring costly, time-intensive data collection[47,48]. Indeed, we hypothesized and confirmed that expanding LLM knowledge extraction to include more variables enhances risk prediction capabilities. Posing questions to an LLM for surrogate modeling using real-world data carries certain risks. If the data happens to be part of the LLM's training corpus, it could lead to an open-book scenario[49], where responses reflect memorized information rather than generalized knowledge. Moreover, this approach raises data privacy concerns, especially when proprietary or sensitive datasets are involved.

Although our study focused on commercial, general-purpose LLMs, the surrogate modeling framework would also be applicable to domain-specific LLMs trained for clinical tasks. In particular, LLMs pre-trained or fine-tuned on clinical data—such as Med-PaLM 2[50], trained on a mixture of curated medical question-answering datasets and other authoritative medical texts, and GatorTron[51], trained on more than 90 billion words from unstructured clinical notes in large-scale EHRs—may encode rich and clinically meaningful associations between patient characteristics and outcomes. Applying our framework to such models could help reveal latent patterns not explicitly described in existing literature, offering new hypotheses for further investigation. Thus, our framework may serve as a useful tool for exploratory analysis in the context of medical foundation models.

LLM abilities emerge through vast data consumption. While surrogate modeling aids in explaining the properties of the LLM knowledge space and contributes to opening the black-box, the phenomena and issues in LLMs ultimately stem from their training data[1,2,11]. The surrogate modeling approach assesses possible inaccuracies and biases in the pre-trained LLM-encoded knowledge, potentially serving as a red-flag indicator. However, addressing these issues also requires research and attention on processes preceding the LLM training stage[1,2,11]. Specifically, studies should also focus on understanding the extent and nature of inaccuracies and biases inherent in the training data, evaluating their impact when included or excluded from training, and identifying strategies to minimize them within training datasets.

Our study had some limitations. First, this study was not intended to provide a comprehensive or systematic evaluation across LLMs, datasets, or modeling strategies. Instead, its primary goal was to introduce a generalizable surrogate modeling framework, demonstrated through selected proof-of-concept experiments on representative LLMs and medical prediction tasks to reveal encoded associations, inaccuracies, and biases. Future work may extend this framework to a broader range of models, topics, prompting strategies, datasets, and statistical approaches for large-scale systematic evaluation. Second, while our study aimed to propose a surrogate modeling framework for LLMs, assessing each coefficient individually for medical appropriateness was beyond our scope. Third, while surrogate modeling can serve as a red-flag indicator for potential inaccuracies or biases, the absence of such findings does not guarantee immunity from them. The lack of observed inaccuracies or biases could stem from flaws in the hypothesis itself— such as improper input variable selection or insufficient statistical modeling. Fourth, to model the relationship between pairs of simulated data and LLM predictions, we employed linear regression for its simplicity and intuitive interpretation. The LLM knowledge was sufficiently explained by modeling with linear regression for the topics of our experiments. However, other topics may require advanced statistical modeling that is more suitable for their specific characteristics, which can only be determined through appropriate exploratory data analysis.

In conclusion, we demonstrated that surrogate modeling is an effective LLM knowledge explanation method for specific hypotheses derived from domain knowledge. The surrogate modeling approach allows for the systematic quantification of potential inaccuracies and biases in LLMs, serving as a red-flag indicator to support the safe and reliable application of these models.

# METHODS

In the main text, we have introduced a surrogate modeling framework to quantitatively explain knowledge encoded in LLMs and have presented experimental results as a proof of concept. Here, we provide a detailed breakdown of the surrogate modeling methodology, methods for LLM implementation, the evaluation process using the UKB dataset, and descriptions of the statistical analyses and ethical considerations.

## Methodological breakdown

Our surrogate modeling framework for explaining LLM-encoded knowledge can be summarized as the quantification of LLM input-output pairs for a target hypothesis derived from domain knowledge achieved through extensive prompting across a wide range of simulated scenarios (Fig. 1). It consists of three stages (Fig. 2).

### Stage 1: Simulated dataset generation

The first stage involves generating a comprehensive and extensive simulated dataset for the target hypothesis, comprising four steps: (1) Outcome selection, (2) Input variable selection, (3) Probability distribution selection, and (4) Random sampling.

1. **Outcome selection**: This stage begins by selecting a topic for LLM knowledge elicitation. For example, in medical prediction contexts, this could include topics such as CVD, stroke or cancer risk prediction, or estimation of GFR.

2. **Input variable selection**: Input variables (e.g., risk factors) relevant to the chosen outcome are then selected. The choice of input variables related to the selected outcome is guided by the target hypothesis derived from prior domain knowledge.
3. **Probability distribution selection**: For each input variable, probability distributions (e.g., normal distribution, lognormal distribution, etc.) are assigned by the user. The choice of probability distributions for each input variable is also guided by the target hypothesis derived from prior domain knowledge. These distributions must be set sufficiently wide to enable comprehensive elicitation of LLM knowledge. For example, in medical prediction contexts, the distributions should be designed to capture a broad range of physiological states and characteristics, reflecting the diversity observed in well-recruited real-world cohorts.
4. **Random sampling**: Extensive random sampling is performed for each variable from the predefined distributions to generate the simulated dataset.

Users can define the target hypothesis derived from domain knowledge for LLM surrogate modeling and specify inputs to generate simulated datasets accordingly. This user-driven approach to the simulated dataset creation offers flexibility and easy re-implementation.

**Stage 2: Obtaining LLM responses**

The second stage involves extensively obtaining LLM responses for a broad and diverse spectrum of simulated scenarios generated in the first stage as follows. The simulated dataset is converted into sentence form and prompted to the LLM to obtain LLM responses (Supplementary Table S2). This process yields pairs of simulated data and LLM responses. To extract the desired outcome from these responses, techniques such as employing carefully designed prompts to ensure consistent formatting in the LLM responses and using regular expressions for extraction can be utilized.

**Stage 3: Statistical modeling**

The third stage involves applying statistical modeling (e.g., linear regression, log-transformed linear regression, polynomial regression, etc) to the pairs of simulated data and LLM responses to quantify and generalize LLM-encoded knowledge into a single statistical formula, making it explainable. The statistical formula (i.e., the surrogate model) reflects the degree to which these models “perceive” each input variable in relation to the desired outcome.

## LLM implementation

Three foundation LLMs were used in a zero-shot setting: GPT-4 (gpt-4-0613) via the OpenAI API[22], Claude-3.5-Sonnet (claude-3-5-sonnet-20241022) via the Anthropic API[23], and Gemini-1.5-Pro (gemini-1.5-pro-002) via the Google Cloud API[24] within a Python environment. Initially, we conducted experiments to determine the optimal temperature setting—a measure of "creative freedom"—which was found to be 0.0 (Supplementary Method S3 and Supplementary Table S14). We used regular expressions to extract the LLM-predicted risk scores. All responses were obtained through separate API calls to ensure response independence.

## Evaluation using the UK Biobank dataset

The UKB cohort represents a sample of the general population of the United Kingdom[25]. Since its inception in 2006, the UKB has gathered longitudinal health and genetic data from over 500,000 participants aged 40-69 at enrollment, with a lengthy follow-up period. Variables extracted from the UKB are listed in Supplementary Table S15.

In Experiment 1, the predictive performance of the surrogate prediction model for MACE was assessed using C-index and compared to traditional clinical prediction models (PCE and FRS)[17,25]. MACE was defined as a combined outcome of myocardial infarction or ischemic stroke, identified using International Classification of Diseases, $10^{th}$ revision (ICD-10) codes I21-I25 for fatal or non-fatal myocardial infarction and I63-I64 for ischemic stroke. Individuals with a history of MACE before the assessment date were excluded. PCE risk scores were classified as low, moderate, and high-risk based on thresholds of 7.5% and 20%, consistent with current cholesterol management guidelines that recommend medium-intensity statins for moderate or high-risk[18]. Among 357,113 individuals analyzed from the UKB (Supplementary Fig. S2), 22,866 (6.4%) experienced MACEs within 10 years (Supplementary Table S4). 215,844 (60.4%), 114,911 (32.2%), and 26,358 (7.4%) individuals were classified as low, moderate, and high-risk by the PCE, respectively. Higher-risk groups were characterized by older age, a higher proportion of males, and a greater prevalence of risk factors—including diabetes, dyslipidemia, CKD, AF, and FHCVD—as well as a higher incidence of 10-year MACE and higher surrogate prediction model scores. To quantify the feature importance of each variable, we applied the SHAP method[27], a model-agnostic tool that provides both global and local interpretability, to the surrogate prediction model.

In Experiment 3 (Supplementary Result S1), the predictive performance of the surrogate prediction model for stroke was evaluated using the C-index and compared with that of the Framingham Stroke Risk Profile (FSRP)[52]. Stroke was defined using ICD-10 codes I60-I64. Individuals with a history of stroke before the assessment date were excluded.

## Ethical statement

This study used data from the UK Biobank under approved application number 85037. The UK Biobank has received ethical approval from the North West Multi-centre Research Ethics Committee as a Research Tissue Bank (REC reference number 21/NW/0157),[25] which covers the ethical approval for registered secondary analyses of UK Biobank data. Under this approval, no additional ethical approval was required for the present study. All participants provided written informed consent to the UK Biobank. The study was conducted in accordance with the UK Biobank guidelines and regulations for researchers and with the principles of the Declaration of Helsinki.

## Statistical analysis

For Harrell's C-index and NRI, 95% CIs were determined using 2,000 bootstrap runs (resampling with replacement), with the 2.5th and 97.5th percentiles reported. Normality of continuous data was assessed via the Shapiro–Wilk test. Normally distributed variables were compared using analysis of variance (ANOVA), while non-normally distributed data were compared using the Kruskal–Wallis test. Categorical variables were compared using the Chi-square test. Model fit was assessed using the R-squared value, with overall model significance evaluated via an F-test. Statistical significance was set at $P < 0.05$ for all analyses.

# FIGURES

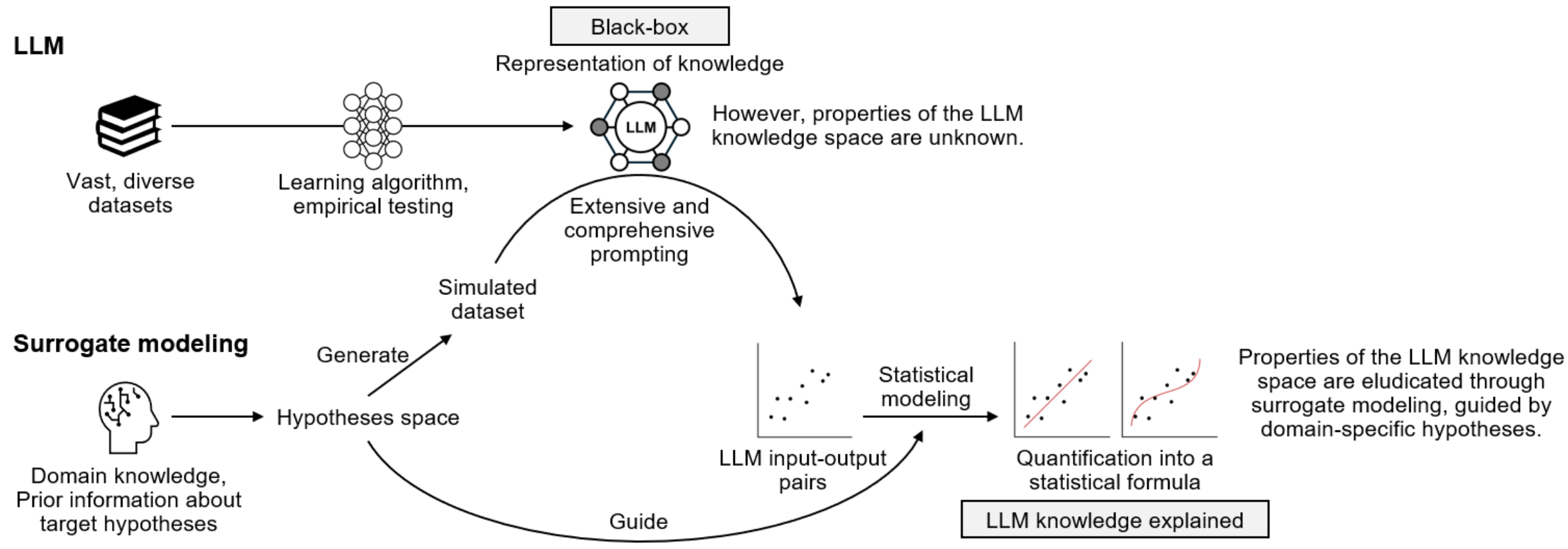

**Fig. 1. Surrogate modeling for LLM black box interpretability.** LLMs are trained using vast and diverse datasets through complex learning algorithms and empirical testing, resulting in a black-box model that encodes extensive hypotheses (knowledge). However, the nature of these hypotheses and their internal structure remain largely opaque. To investigate this black-box model, we propose applying surrogate modeling to quantitatively explain the knowledge encoded in LLMs for specific hypotheses. A simulated dataset is generated, guided by a target hypothesis derived from prior domain knowledge. This dataset is then used to extensively and systematically prompt the LLM, eliciting its knowledge in the form of input-output pairs. These input-output pairs are subsequently statistically modeled, guided by the target hypothesis derived from domain knowledge. The resulting statistical formula reflects the degree to which the LLM "perceives" each input variable in relation to the desired outcome. In short, the properties of the LLM knowledge space are elucidated through surrogate modeling, guided by domain-specific hypotheses.

LLM: large language model.

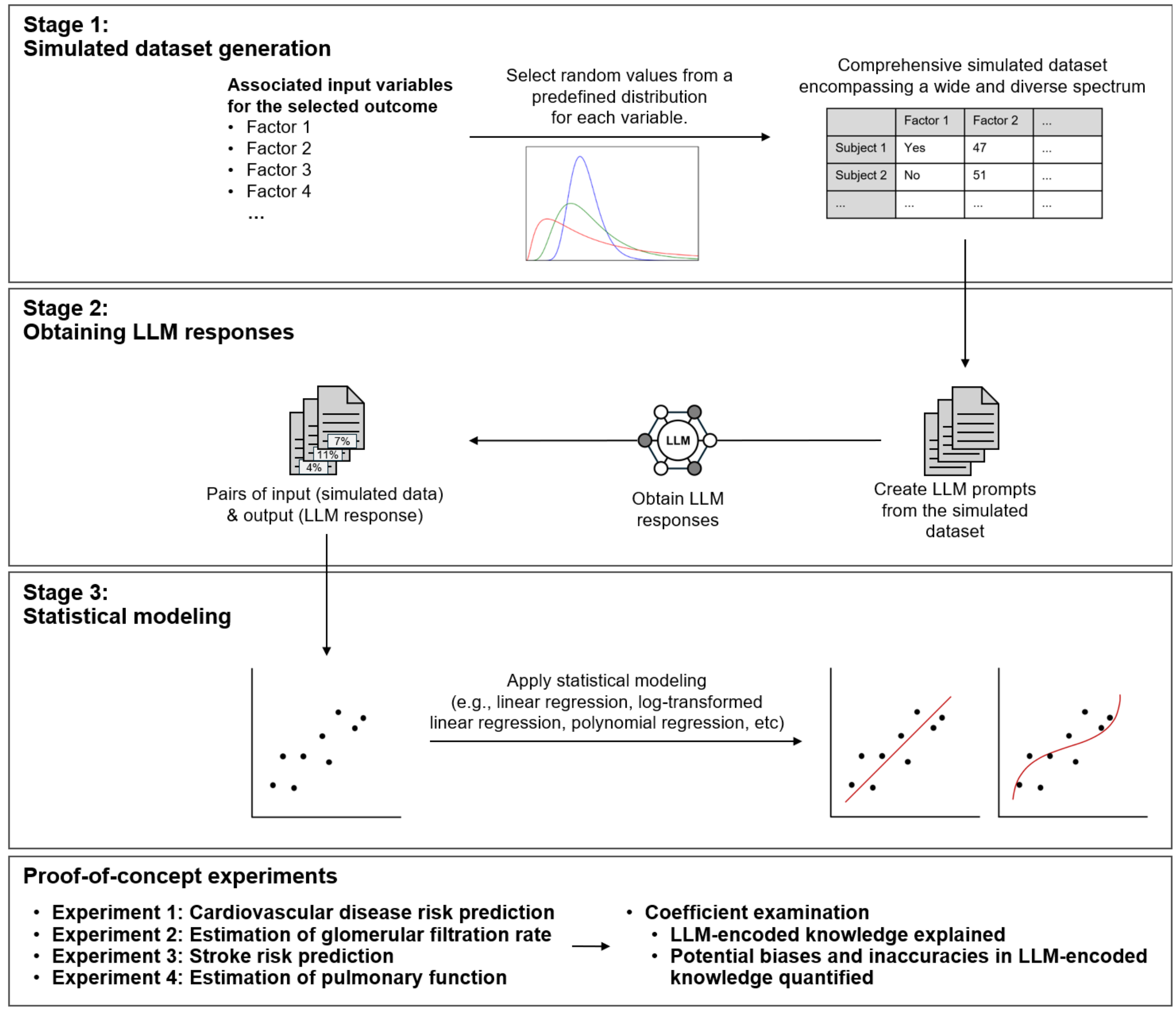

**Fig. 2. Overview of the surrogate modeling framework for quantitative explanation of LLM-encoded knowledge.** Our surrogate modeling framework quantifies LLM input-output pairs for a target hypothesis derived from domain knowledge, using extensive prompting across diverse simulated scenarios (Fig. 1). It comprises three stages: (1) generating a simulated dataset by selecting outcomes, input variables, and probability distributions guided by the target hypothesis; (2) obtaining LLM responses for the simulated scenarios; and (3) applying statistical modeling to translate LLM responses into a single explainable statistical formula. As a proof of concept, we present medical prediction experiments demonstrating how surrogate modeling elucidates LLM-encoded knowledge, revealing biases and inaccuracies.

LLM: large language model, CVD: cardiovascular disease, GFR: glomerular filtration rate.

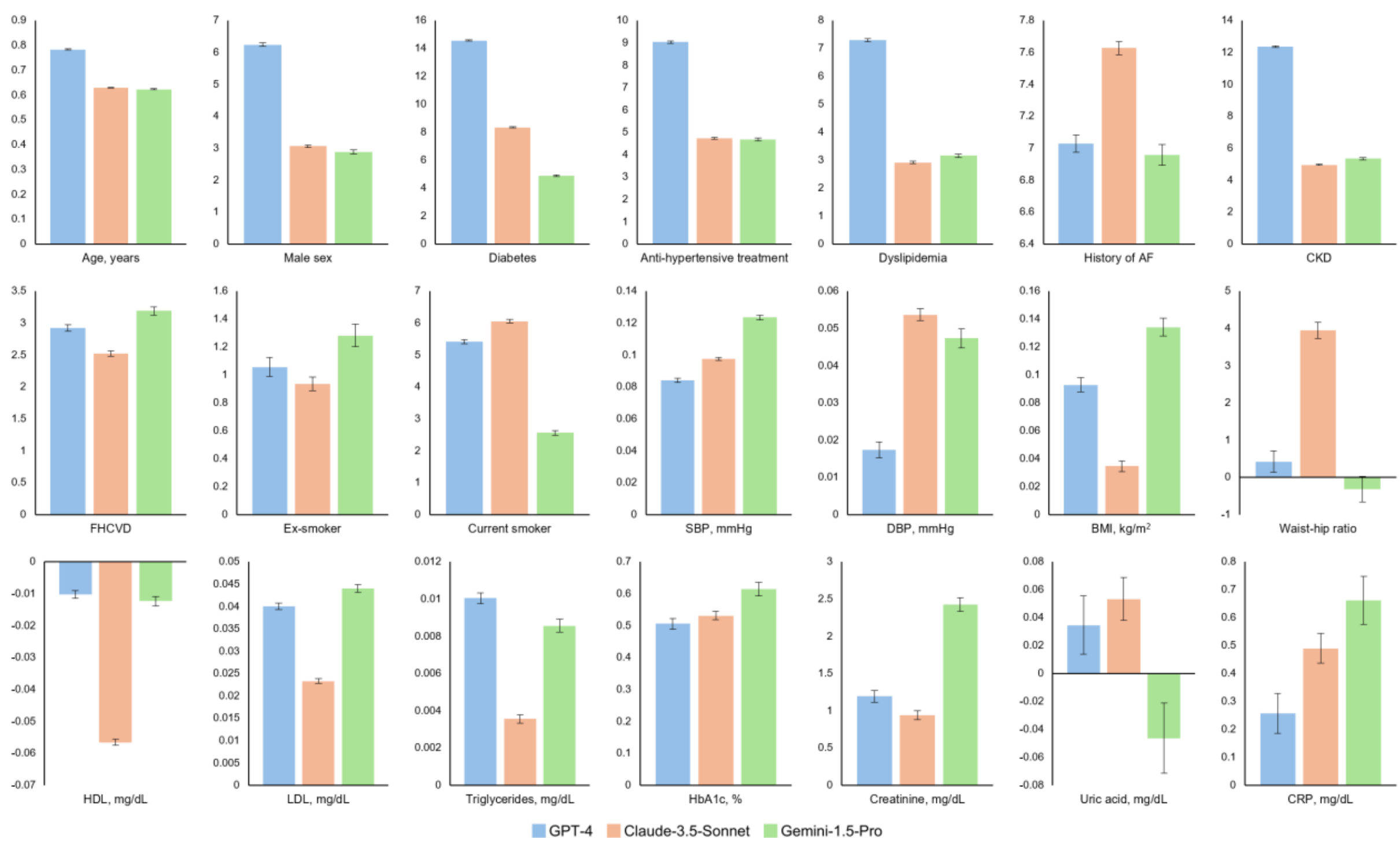

**Fig. 3. Regression coefficient bar plots (Experiment 1).** LLM-encoded knowledge is explained through the examination of the coefficients. The error bars represent the 95% confidence intervals.

LLM: large language model, AF: atrial fibrillation, CKD: chronic kidney disease, FHCVD: family history of cardiovascular disease in first degree relatives, SBP: systolic blood pressure, DBP: diastolic blood pressure, BMI: body mass index, HDL: high-density lipoprotein, LDL: low-density lipoprotein, HbA1c: hemoglobin A1c, CRP: C-reactive protein

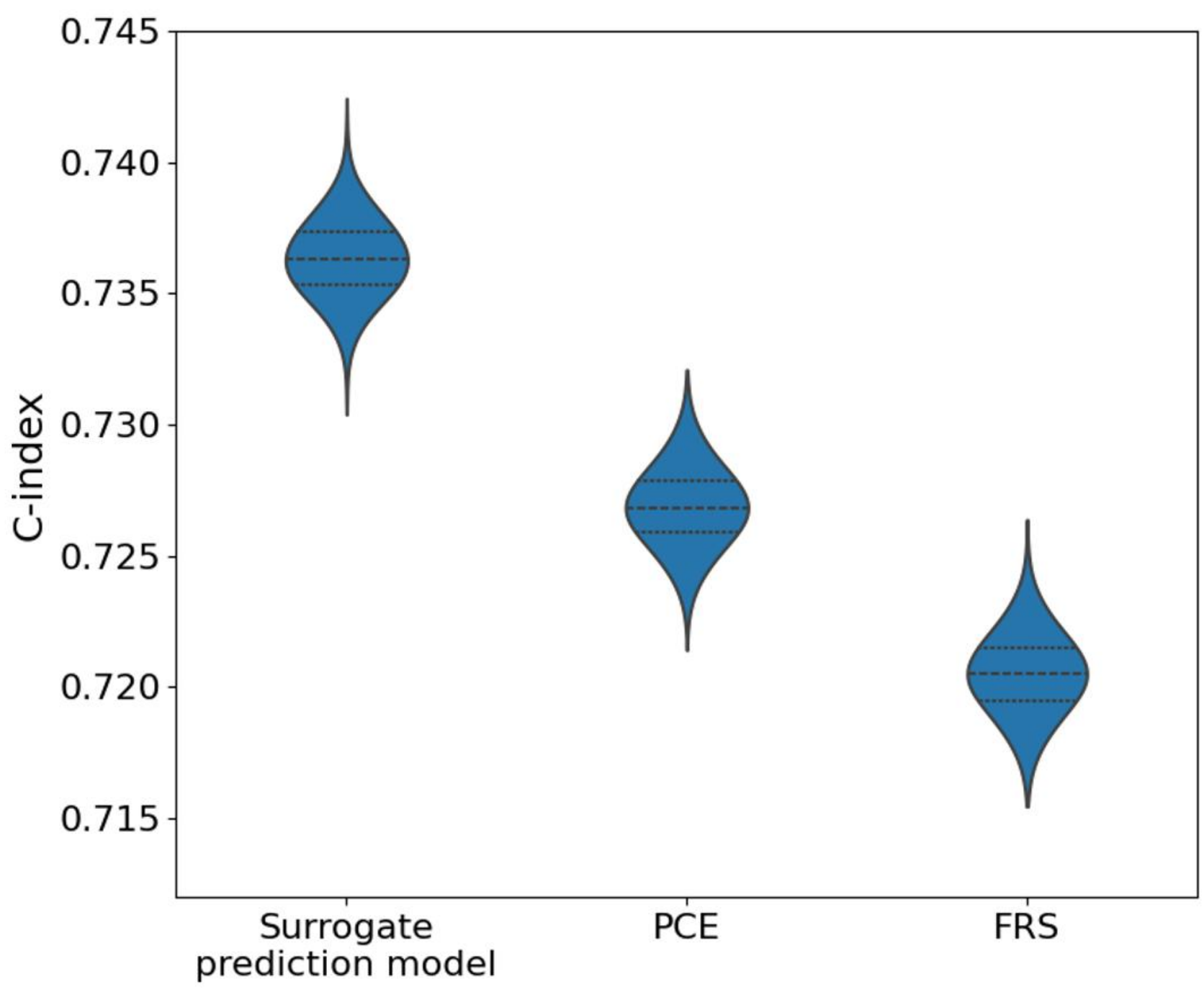

**Fig. 4. Predictive ability of the surrogate prediction model.** Violin plot showing the C-index for MACE from 2000 bootstrap runs.

MACE: major adverse cardiovascular event, PCE: Pooled Cohort Equations, FRS: Framingham Risk Score.

**(A) LLMs, on their own, yield inconsistent responses and cannot explain variable contributions.**

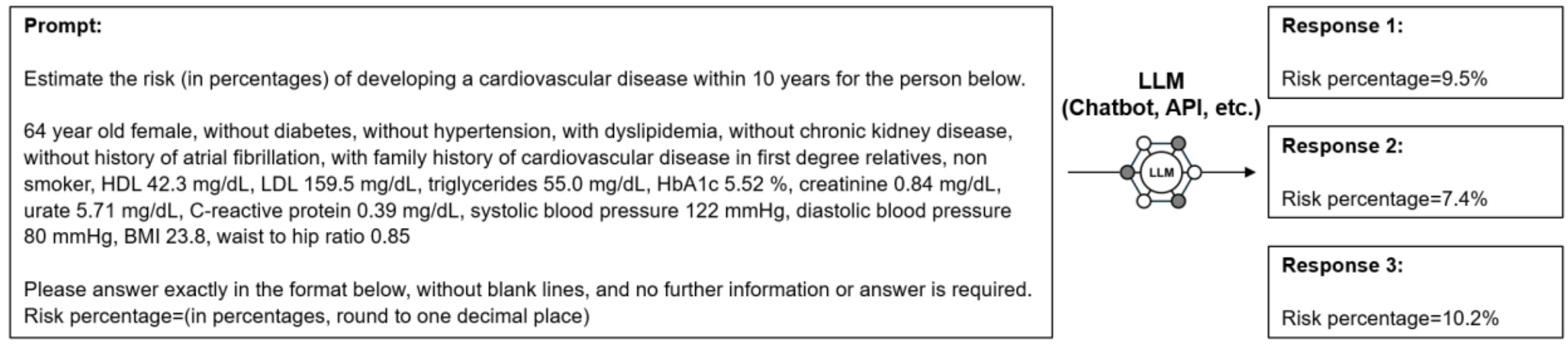

**(B) Our surrogate modeling framework resolves inconsistency and explains variable contributions.**

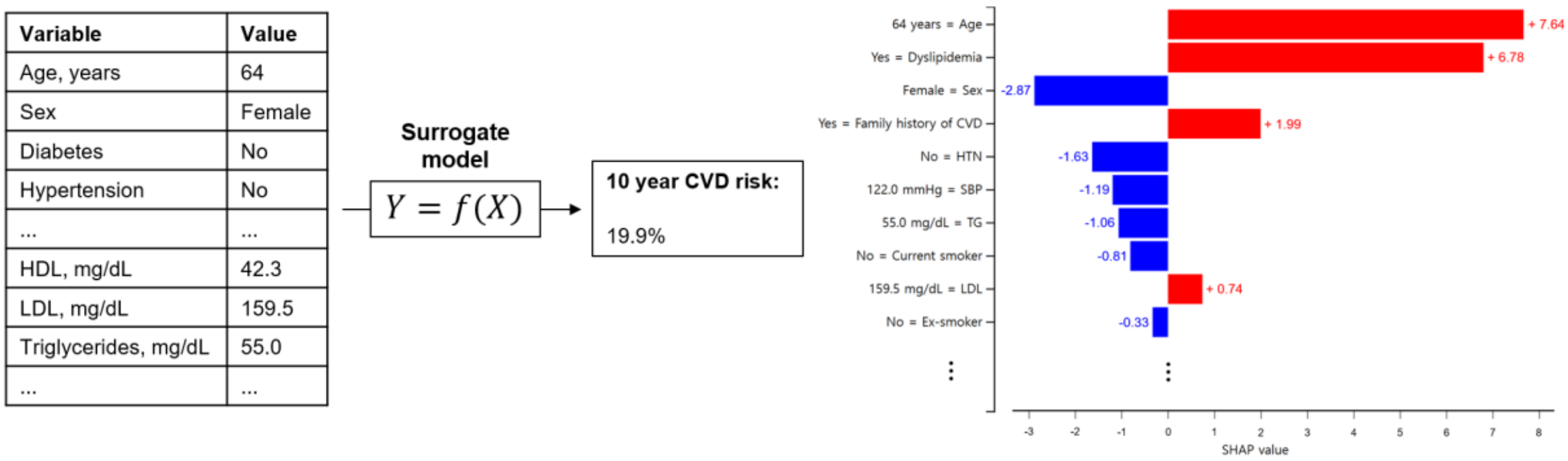

**Fig. 5. Surrogate modeling resolves inconsistency and explains variable contributions.** (A) LLMs inherently produce different responses to the same input each time and cannot quantify the contribution of individual variables to the output. (B) The surrogate modeling framework resolves inconsistency through a generalized statistical formula of LLM-encoded knowledge and quantifies the contribution of each variable to the output using SHAP values.

LLM: large language model, SHAP: SHapley Additive exPlanations, HDL: high-density lipoprotein, LDL: low-density lipoprotein, HbA1c: hemoglobin A1c, BMI: body mass index, CVD: cardiovascular disease, HTN: hypertension, TG: triglycerides.

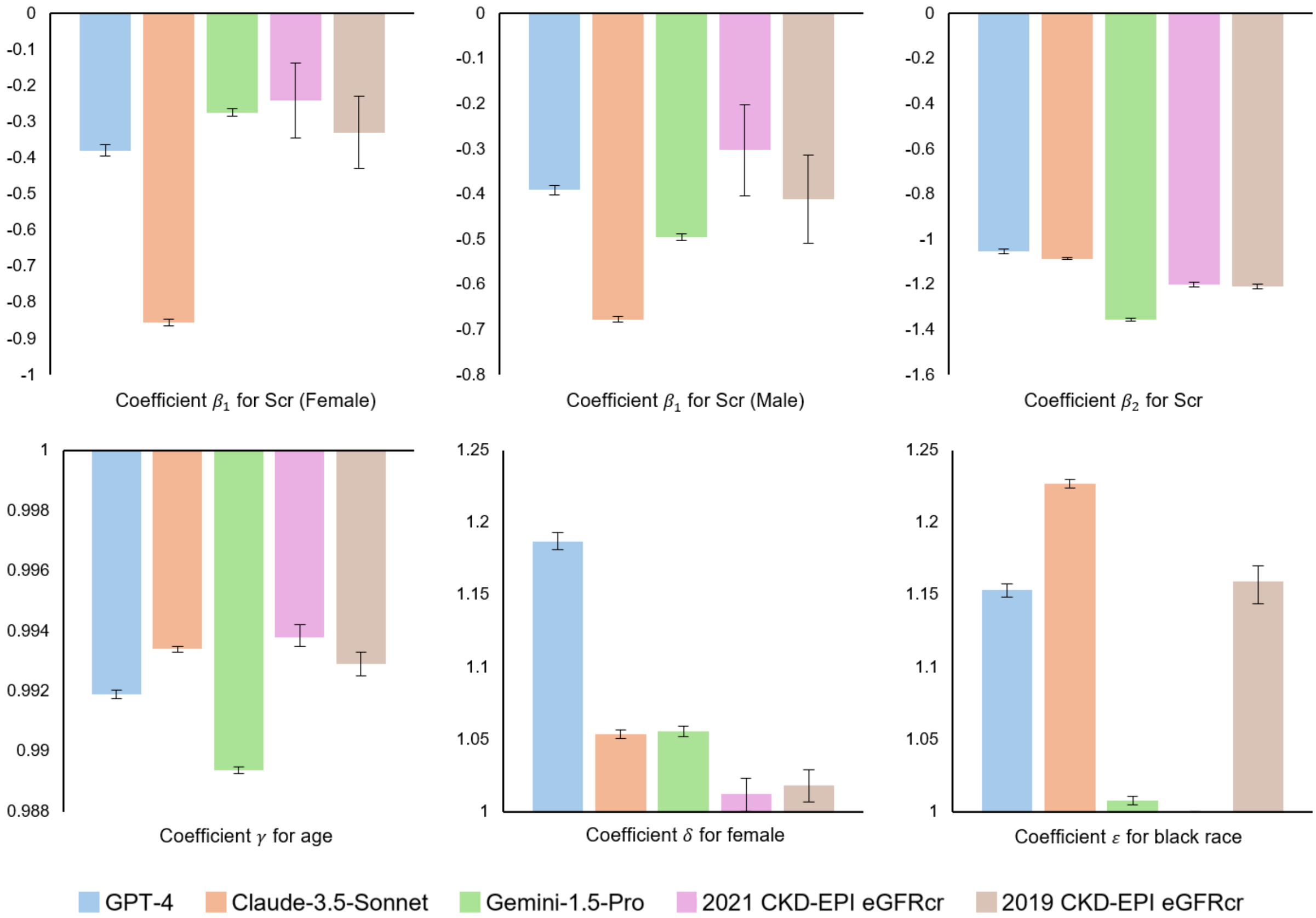

**Fig. 6. Regression coefficient bar plots (Experiment 2).** LLM-encoded knowledge is explained through the examination of the coefficients. The error bars represent the 95% confidence intervals.

LLM: large language model, Scr: serum creatinine, CKD-EPI: Chronic Kidney Disease Epidemiology Collaboration, eGFRcr: glomerular filtration rate estimated with the use of creatinine

## DATA AVAILABILITY

UK Biobank data are available to researchers at http://www.ukbiobank.ac.uk/using-the-resource/. The pairs of simulated data and LLM predictions, along with our code, are available at https://github.com/CMI-Laboratory/LLM_knowledge_explanation/. An online tool for our surrogate prediction model is available at https://llm-surrogate.com/.

## ACKNOWLEDGEMENTS AND FUNDING

This research was supported by a grant of the Korea Health Technology R&D Project through the Korea Health Industry Development Institute (KHIDI), funded by the Ministry of Health & Welfare, Republic of Korea (grant number: RS-2022-KH125153). This work was supported by the National Research Foundation of Korea(NRF) grant funded by the Korea government (MSIT) (RS-2023-00276320). L.A.C. is funded by the National Institute of Health through R01 EB017205, DS-I Africa U54 TW012043-01 and Bridge2AI OT2OD032701, and the National Science Foundation through ITEST #2148451. This study used data from UK Biobank (application number: 85037). The funders had no role in the design and conduct of the study; collection, management, analysis, and interpretation of the data; preparation, review, or approval of the manuscript; and decision to submit the manuscript for publication. This study was conducted while C.H. was affiliated with the Department of Biomedical Systems Informatics, Yonsei University College of Medicine, Seoul, Korea. He is currently affiliated with the Medical Big Data Research Center, Seoul National University Medical Research Center, Seoul National University College of Medicine, Seoul, Korea.

## AUTHOR CONTRIBUTIONS

C.H. and S.K. conceived and designed the overall study. C.H., S.K. and D.W.K. designed and developed the detailed methodological approach for this study. D.W.K. extracted the initial data. C.H. analyzed the data, implemented the Python and R codes, and performed statistical analyses. J.K. provided advice regarding the statistical analyses. C.H., S.K. and D.W.K. interpreted the results and drafted the manuscript. L.A.C., D.Y. and S.B. critically revised the manuscript and improved the study conception and design. D.Y. and S.B. supervised this study. D.Y. and L.A.C. obtained funding. All authors revised and approved the final version of the manuscript.

## COMPETING INTERESTS

The authors declare no competing interests.

# Supplementary Information

## Supplementary methods

## Supplementary results

## Supplementary references

## Supplementary figures

## Supplementary tables

# Supplementary methods

### Supplementary Method S1. Variable selection (Experiment 1)

Experiment 1 started with selecting established cardiovascular (CVD) risk factors, in addition to those included in the Pooled Cohort Equation (PCE)[1]. The PCE encompasses age, sex, total cholesterol, high-density lipoprotein (HDL) cholesterol, systolic blood pressure (SBP), diabetes status, anti-hypertensive treatment, smoking status, and race[1]. After extensive literature review, we selected additional well-established CVD risk factors. These additional factors can be categorized into three groups: medical history (including dyslipidemia[2,3], history of atrial fibrillation[4,5], chronic kidney disease[6,7], and family history of CVD in first degree relatives)[8,9], physical measurements and examinations (including diastolic blood pressure[10], body mass index[11,12], and waist–hip ratio[12,13]), and routinely measured biochemical markers (including low-density lipoprotein [LDL] cholesterol[14,15], triglycerides[16], hemoglobin A1c[17], creatinine[18,19], uric acid[20,21], and C-reactive protein[22,23]).

### Supplementary Method S2. Lognormal distribution

The median value and either the 99th or 1st percentile value were set to determine the lognormal distribution uniquely for each variable while ensuring an adequate reflection of both physiologically normal and abnormal states. Subsequently, the values exceeding the 99.9th percentile or falling below the 0.1st percentile were excluded from the distribution to eliminate extreme outliers. Random instances were independently extracted from the resulting distribution for each variable to construct the simulated dataset. The density plots of the values extracted for each variable are shown in Supplementary Fig. S1. Following the extraction of the HDL cholesterol, LDL cholesterol, and triglycerides using this method, the total cholesterol was calculated post-hoc using the Friedewald equation (total cholesterol = HDL + LDL + triglycerides/5, all units in mg/dL)[24]. Total cholesterol was excluded from the input variables in the linear regression model, as it was calculated by simply summing HDL cholesterol, LDL cholesterol, and triglycerides.

### Supplementary Method S3. Temperature experiments

Initially, we conducted experiments to find the optimal temperature parameter for GPT-4, which serves as a measure of “creative freedom”[25]. We incrementally adjusted the temperature from 0.0 to 1.0. We assessed the predictive performance of the constructed surrogate prediction model for major adverse cardiac events (MACEs) through the C-index, the average score of the surrogate prediction model in the United Kingdom Biobank (UKB), and response validity. Response validity was defined as the ratio of responses that adhered to the prompt request, specifically those that provided answers in the exact form of “Risk percentage = 5.2%”, thereby enabling the extraction of the risk percentage via regular expression. The results of the temperature experiments are shown in Supplementary Table S14. Although the differences were minimal, the C-index for the surrogate prediction model was the highest at a temperature of 0.0. As the temperature increased, the average score of the surrogate prediction model in the UKB also increased, diverging from the actual 10-year MACE incidence, which was 6.40% in the UKB data, and the response validity decreased. In conclusion, we identified a temperature of 0.0 as the optimal setting. Similarly, we set the lowest temperature (0.0) for experiments with both Claude-3.5-Sonnet and Gemini-1.5-Pro.

## Supplementary results

**Supplementary Result S1. Experiment 3: Stroke risk prediction**

In the third experiment, we focused on stroke risk prediction. We followed the same principles as in the first experiment. We selected risk factors related to stroke and generated a simulated dataset of 20,000 samples through random sampling, using predefined distributions for each variable. We then structured the simulated dataset into sentence form to create LLM prompts aimed at eliciting 10-year stroke risk predictions (Supplementary Table S2). For each sample, we collected 10 responses from each of the three LLMs. Finally, we developed a surrogate prediction model through linear regression. Afterward, we conducted a coefficient examination to derive a generalized explanation of the LLM-encoded knowledge. We also compared the surrogate prediction model with the Revised Framingham Stroke Risk Profile (FSRP) using UKB data[26].

A total of 19 well-established stroke risk factors were selected (Supplementary Table S10, Supplementary Fig. S5). The linear regression coefficients quantify the LLM-encoded knowledge for stroke risk. The disparity in coefficients between the LLM models reflects the difference in the knowledge each model possesses, or the degree to which each LLM "thinks" about each variable in relation to stroke risk prediction.

Among 370,149 individuals analyzed from the UKB (Supplementary Fig. S2), 8612 (2.3%) experienced stroke within 10 years. Differences were observed in the distribution of predicted risks across the FSRP, and surrogate models for GPT-4, Claude-3.5-Sonnet, and Gemini-1.5-Pro (Supplementary Table S11). The surrogate prediction model for GPT-4 had a C-index of 0.711 (95% CI: 0.706–0.716), and the FSRP had 0.712 (95% CI: 0.707–0.717), showing no significant difference (-0.001, 95% CI: -0.003 – 0.001). The surrogate prediction model for Claude-3.5-Sonnet had a C-index of 0.713 (95% CI: 0.707–0.718), with no significant difference from the FSRP (0.001, 95% CI: -0.002 – 0.003). The surrogate prediction model for Gemini-1.5-Pro had a C-index of 0.712 (95% CI: 0.706–0.717), also showing no significant difference from the FSRP (-0.001, 95% CI: -0.003 – 0.002).

**Supplementary Result S2. Experiment 4: Estimation of pulmonary function**

In the fourth experiment, we focused on estimating pulmonary function. Spirometry is a common test that measures lung function, specifically assessing forced vital capacity (FVC), forced expiratory volume in one second (FEV1), peak expiratory flow rate (PEFR), etc. There are established equations for estimating FVC, FEV1, and PEFR using age, sex, and height[27,28,29]. Historically, these equations included race-based adjustments, but this practice has been scientifically refuted[30,31,32].

Therefore, in "Stage 1-2: Input variable selection", we included age, sex, height, and race (White or Black) as input variables to quantitatively investigate whether any racial bias still persists in the LLM-encoded knowledge. We followed the same principles as in the previous experiments to generate a simulated dataset of 20,000 samples and prompted the LLMs to provide estimations for FVC, FEV1, or PEFR. For each sample, we collected a single response from each of the three LLMs. Subsequently, we performed statistical modeling for each sex as follows:

(1) Linear regression (FVC):

$$FVC\ (men) = \beta_{0,m}^{(1)} + \beta_{1,m}^{(1)} * age + \beta_{2,m}^{(1)} * Male + \beta_{3,m}^{(1)} * Black\ race$$

$$FVC\ (women) = \beta_{0,f}^{(1)} + \beta_{1,f}^{(1)} * age + \beta_{2,f}^{(1)} * Male + \beta_{3,f}^{(1)} * Black\ race$$

(2) Linear regression (FEV1):

$$FEV1\ (men) = \beta_{0,m}^{(2)} + \beta_{1,m}^{(2)} * age + \beta_{2,m}^{(2)} * Male + \beta_{3,m}^{(2)} * Black\ race$$

$$FEV1\ (women) = \beta_{0,f}^{(2)} + \beta_{1,f}^{(2)} * age + \beta_{2,f}^{(2)} * Male + \beta_{3,f}^{(2)} * Black\ race$$

(3) Linear regression (PEFR):

$$PEFR\ (men) = \beta_{0,m}^{(3)} + \beta_{1,m}^{(3)} * age + \beta_{2,m}^{(3)} * Male + \beta_{3,m}^{(3)} * Black\ race$$

$$PEFR\ (women) = \beta_{0,f}^{(3)} + \beta_{1,f}^{(3)} * age + \beta_{2,f}^{(3)} * Male + \beta_{3,f}^{(3)} * Black\ race$$

In the experiment utilizing GPT-4, the coefficient for Black race in estimating FVC was -0.013 L (95% CI: -0.023 – -0.003) for men and -0.162 L (95% CI: -0.168 – -0.156) for women (Supplementary Table S12, Supplementary Fig. S6). This suggests that the LLM estimates FVC values to be 0.013 L lower for Black men and 0.162 L lower for Black women compared to White individuals, after adjusting for other variables. Similarly, racial bias was observed in the LLM-encoded knowledge when estimating FVC, FEV1, and PEFR across all three LLMs.

Differences were observed in the distributions of estimated pulmonary function values (FVC, FEV1 and PEFR) across direct LLM estimations, the corresponding surrogate models, and a reference equation within the simulated dataset (Supplementary Table S13). Direct estimations from each LLM tended to cluster at specific values, while those from the surrogate models exhibited smoother, more continuous distributions (Supplementary Fig. S7).

# Supplementary references

# Supplementary figures

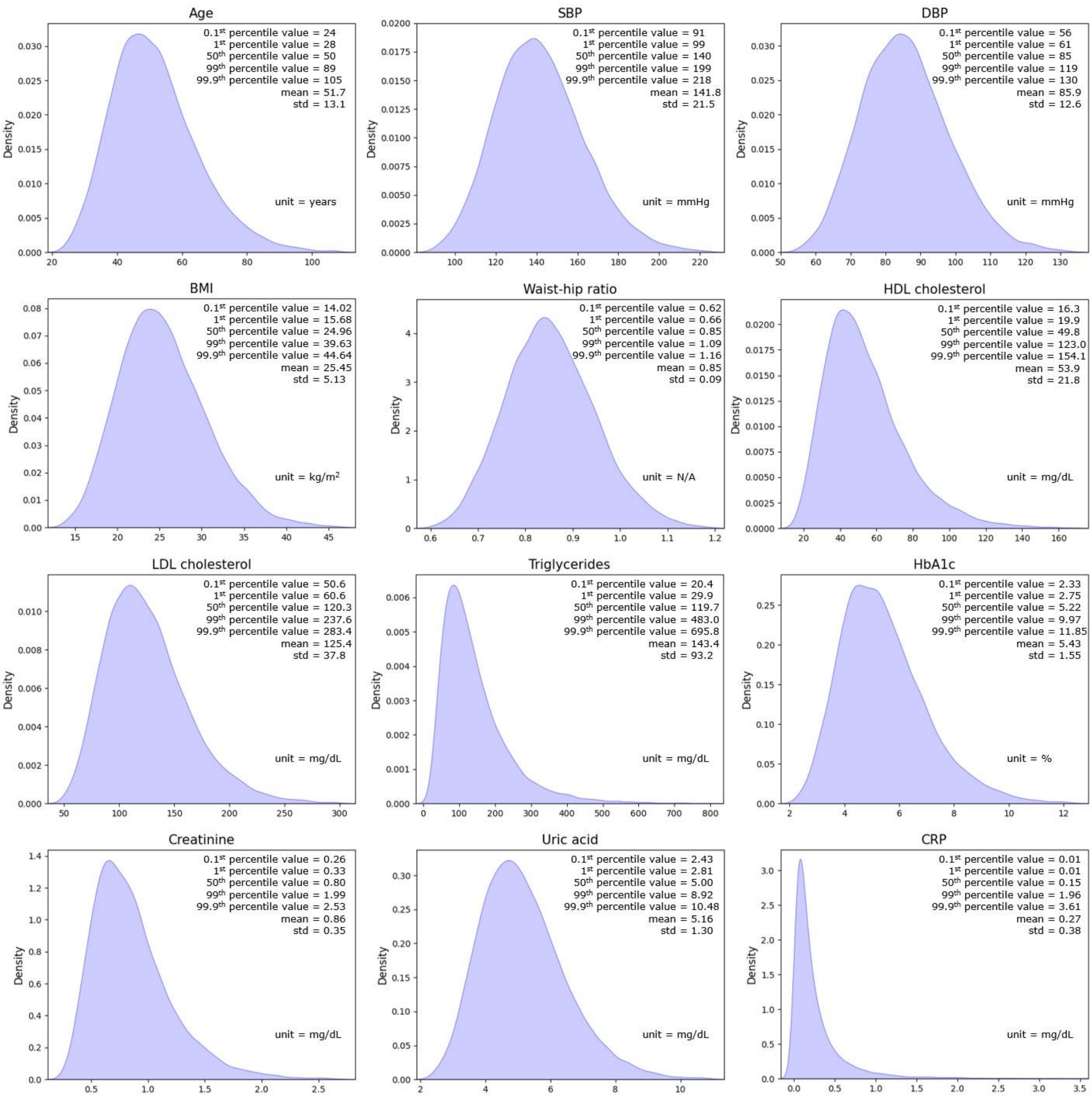

**Supplementary Fig. S1. Density plots for the continuous variables in the simulated dataset (Experiment 1).** We selected a probability distribution for each variable and generated a simulated dataset of 20,000 samples through random sampling. For continuous variables, we used a broad lognormal distribution to reflect both physiologically normal and abnormal states, mimicking a well-recruited real-world cohort. The figure shows the density plots, as well as the 0.1st, 1st, 50th, 99th, and 99.99th percentile values, alongside the mean and standard deviation values for the continuous variables in the simulated dataset. For categorical variables, we allocated equal sample sizes across categories in the simulated dataset.

SBP: systolic blood pressure, DBP: diastolic blood pressure, BMI: body mass index, HDL: high-density lipoprotein, LDL: low-density lipoprotein, HbA1c: hemoglobin A1c, CRP: C-reactive protein. std: standard deviation.

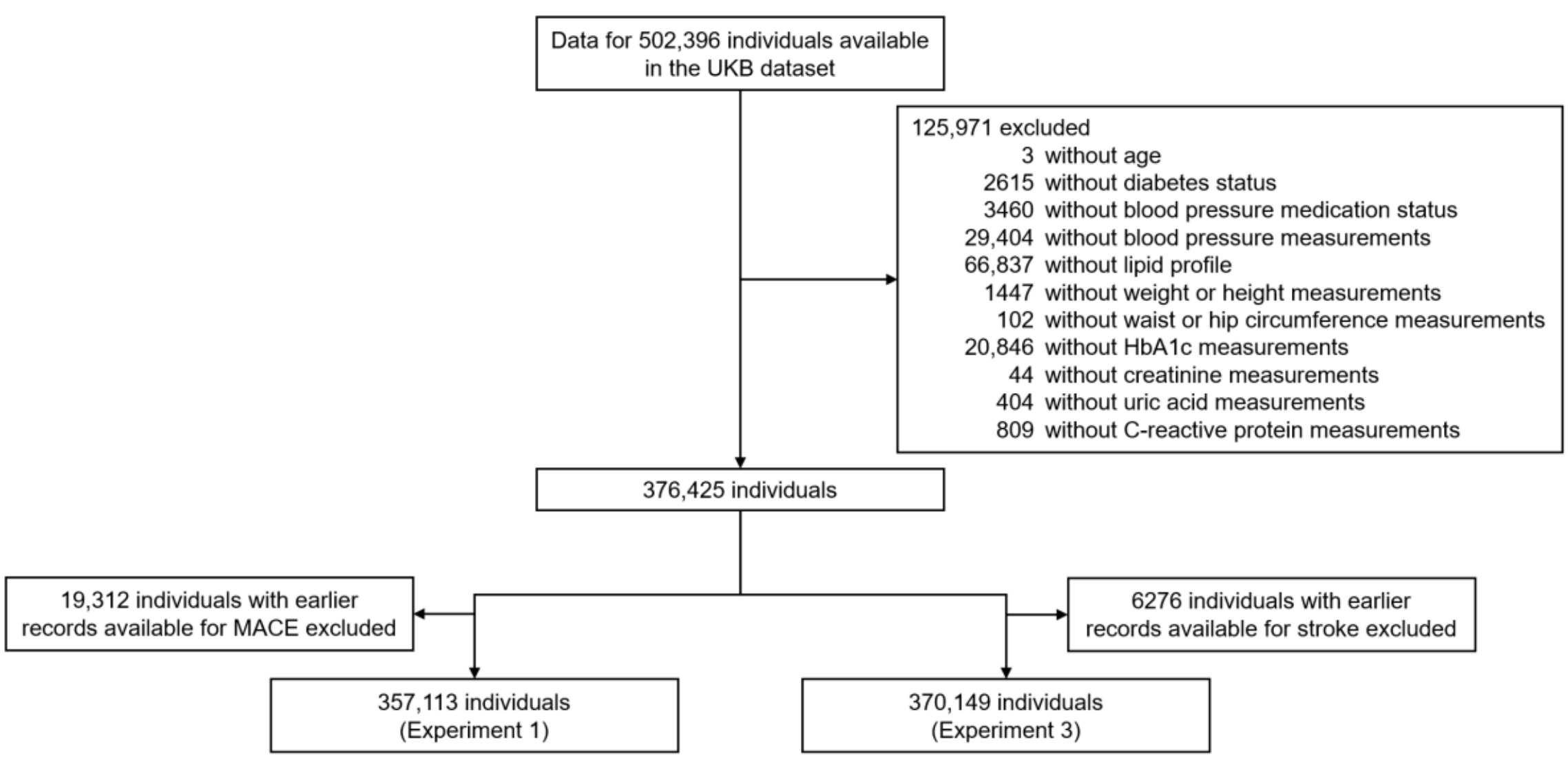

**Supplementary Fig. S2. Selection of UKB study population.**

UKB: United Kingdom Biobank, HbA1c: hemoglobin A1C, MACE: major adverse cardiovascular event.

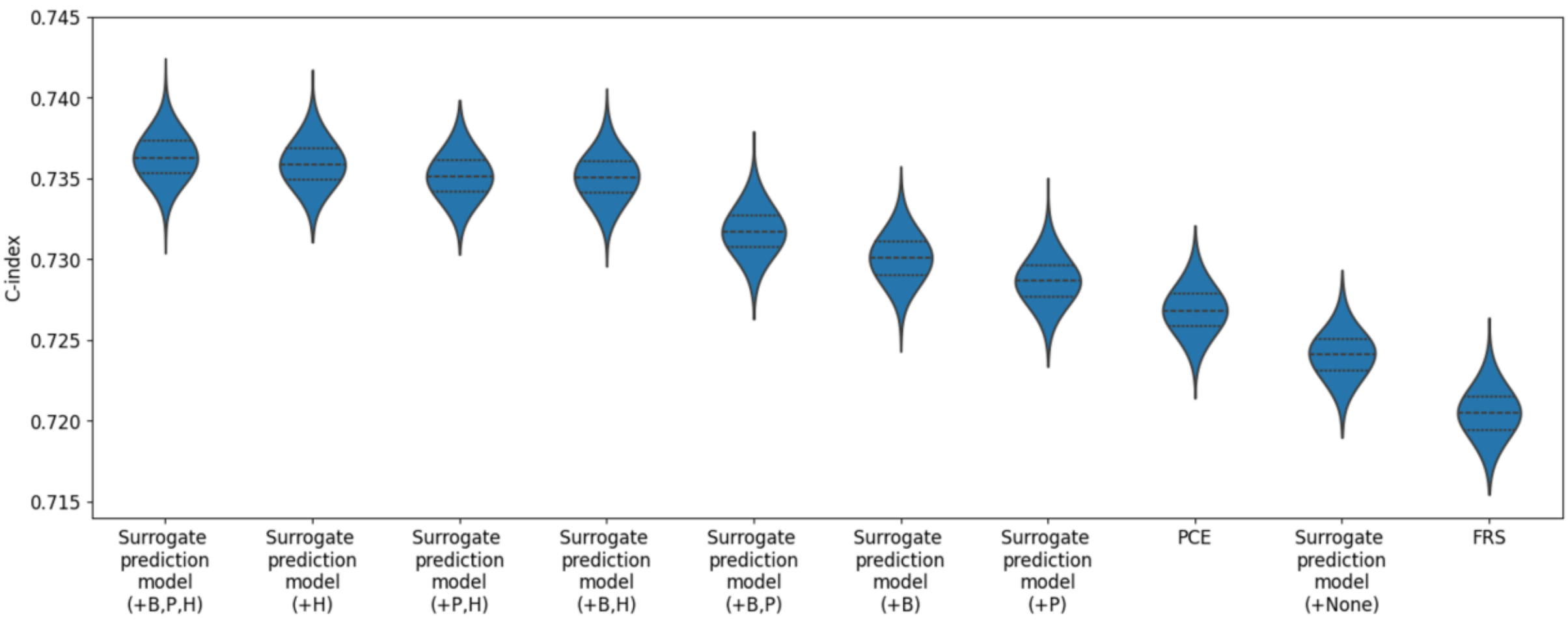

**Supplementary Fig. S3. Predictive ability of the surrogate prediction model for various input variable combinations.** Violin plot showing the C-index for MACE from 2000 bootstrap runs. "Surrogate prediction model (+None)" denotes a baseline using PCE variables for the surrogate prediction model. "+B" adds biochemical markers (LDL, triglycerides, HbA1c, creatinine, uric acid, CRP), "+P" adds physical measurements (DBP, BMI, waist-hip ratio), and "+H" adds medical history (dyslipidemia, AF, CKD, FHCVD).

MACE: major adverse cardiovascular event, PCE: Pooled Cohort Equations, LDL: low-density lipoprotein, HbA1c: hemoglobin A1c, CRP: C-reactive protein, DBP: diastolic blood pressure, BMI: body mass index, AF: atrial fibrillation, CKD: chronic kidney disease, FHCVD: family history of cardiovascular disease in first-degree relatives, FRS: Framingham Risk Score.

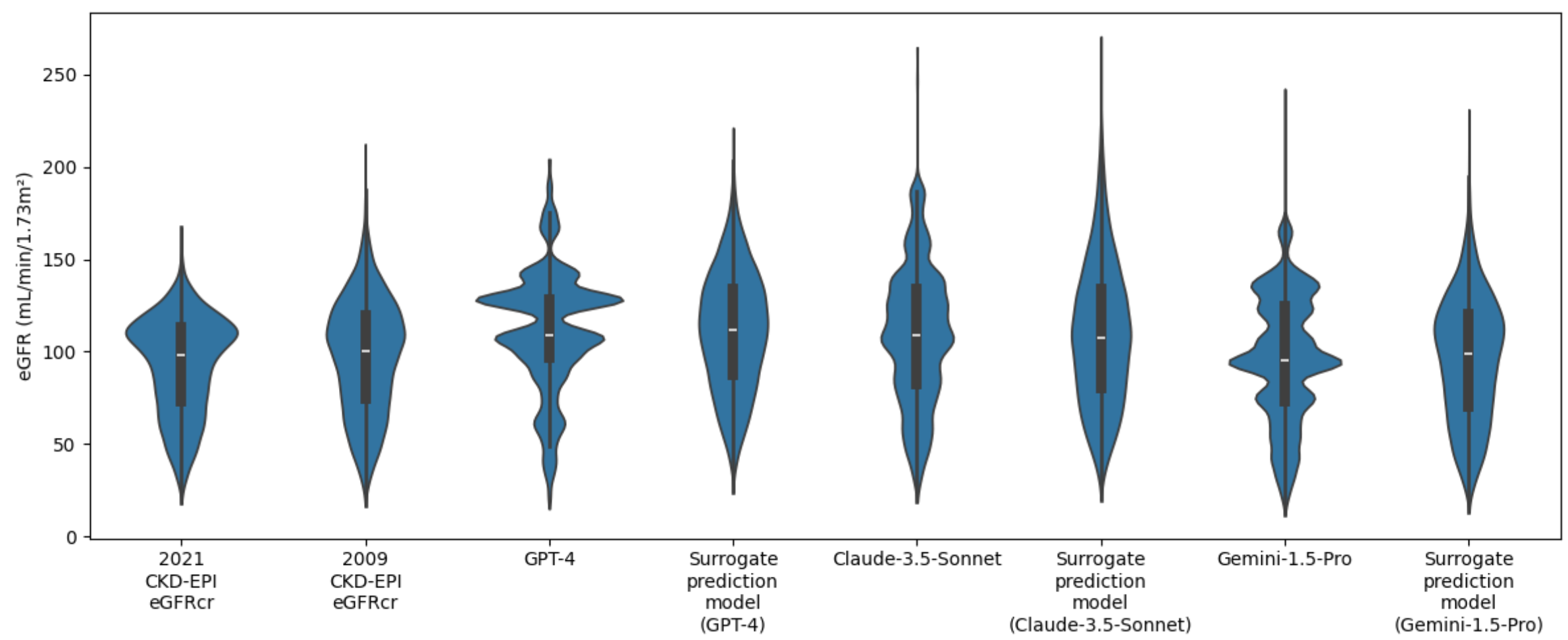

**Supplementary Fig. S4. Violin plots comparing eGFR estimates from CKD-EPI formulas, direct LLM predictions, and their corresponding surrogate models (Experiment 2).**

eGFR: estimated glomerular filtration rate, CKD-EPI: Chronic Kidney Disease Epidemiology Collaboration, LLM: large language model, eGFRcr: glomerular filtration rate estimated with the use of creatinine

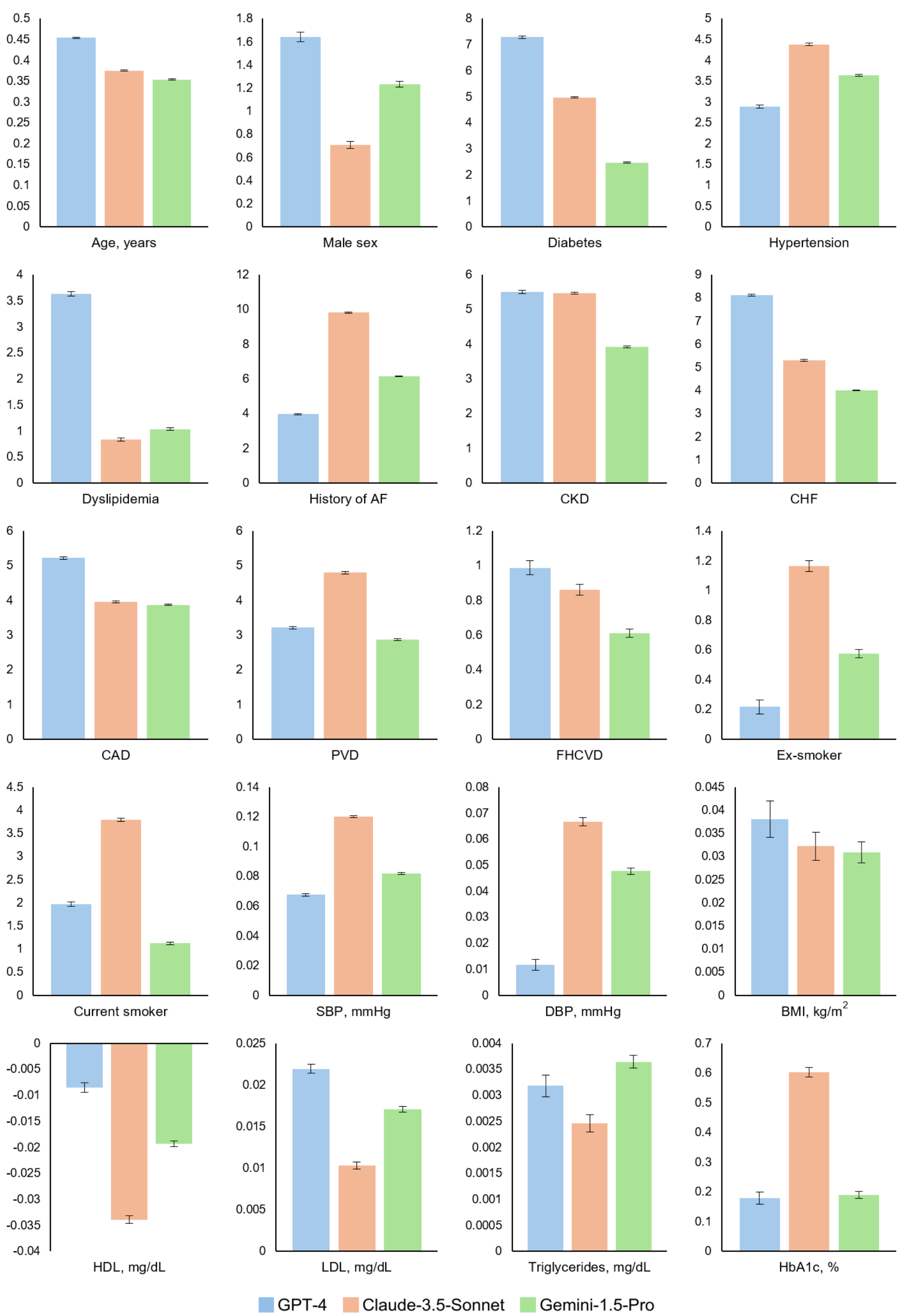

**Supplementary Fig. S5. Regression coefficient bar plots (Experiment 3).** LLM-encoded knowledge is explained through the examination of the coefficients. The error bars represent the 95% confidence intervals.

LLM: large language model, AF: atrial fibrillation, CKD: chronic kidney disease, CHF: congestive heart failure, CAD: coronary artery disease, PVD: peripheral vascular disease, FHCVD: family history of cardiovascular disease in first degree relatives, SBP: systolic blood pressure, DBP: diastolic blood pressure, BMI: body mass index, HDL: high-density lipoprotein, LDL: low-density lipoprotein, HbA1c: hemoglobin A1c.

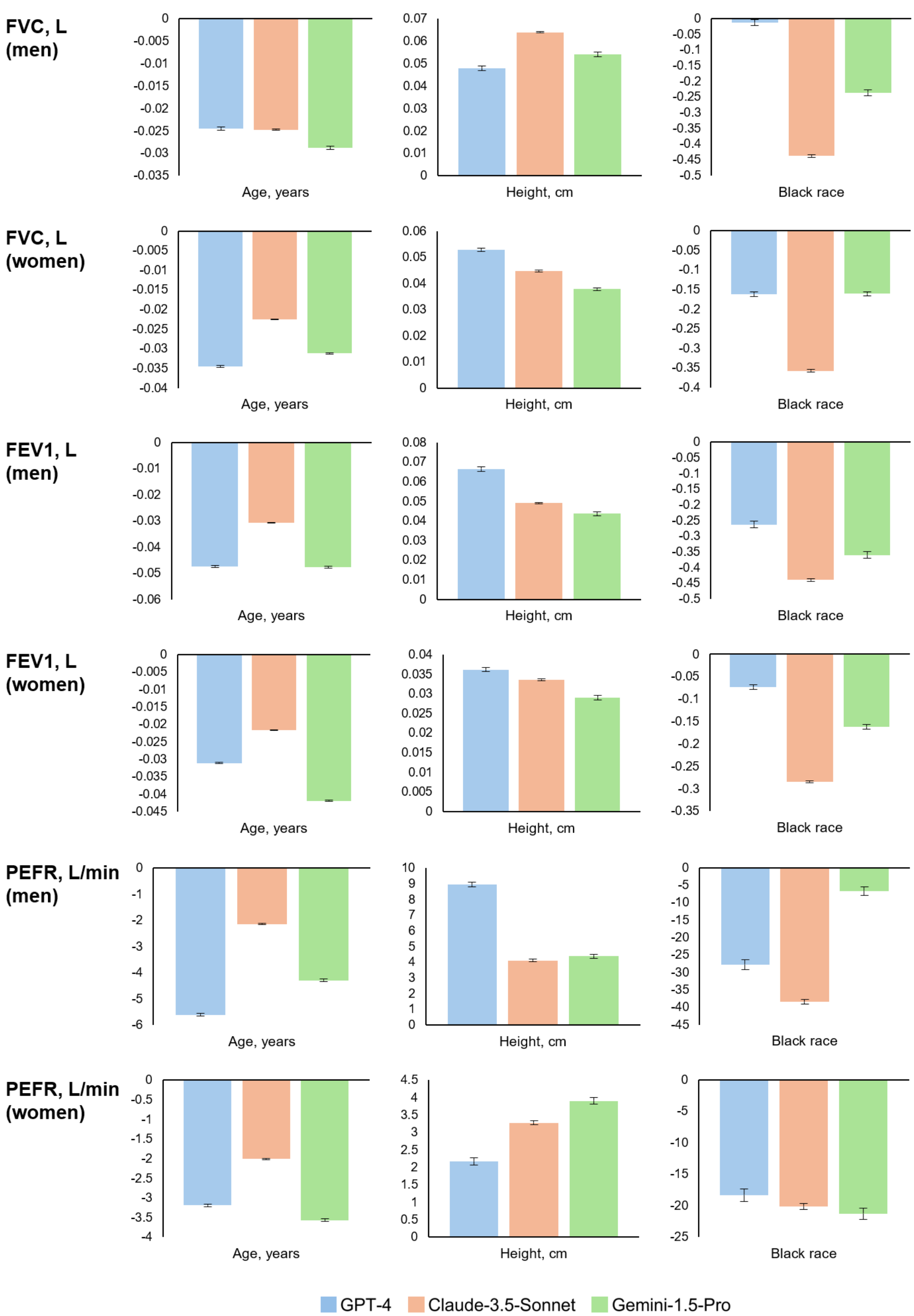

**Supplementary Fig. S6. Regression coefficient bar plots (Experiment 4).** LLM-encoded knowledge is explained through the examination of the coefficients. The error bars represent the 95% confidence intervals.

LLM: large language model, FVC: forced vital capacity, FEV1: forced expiratory volume in one second, PEFR: peak expiratory flow rate.

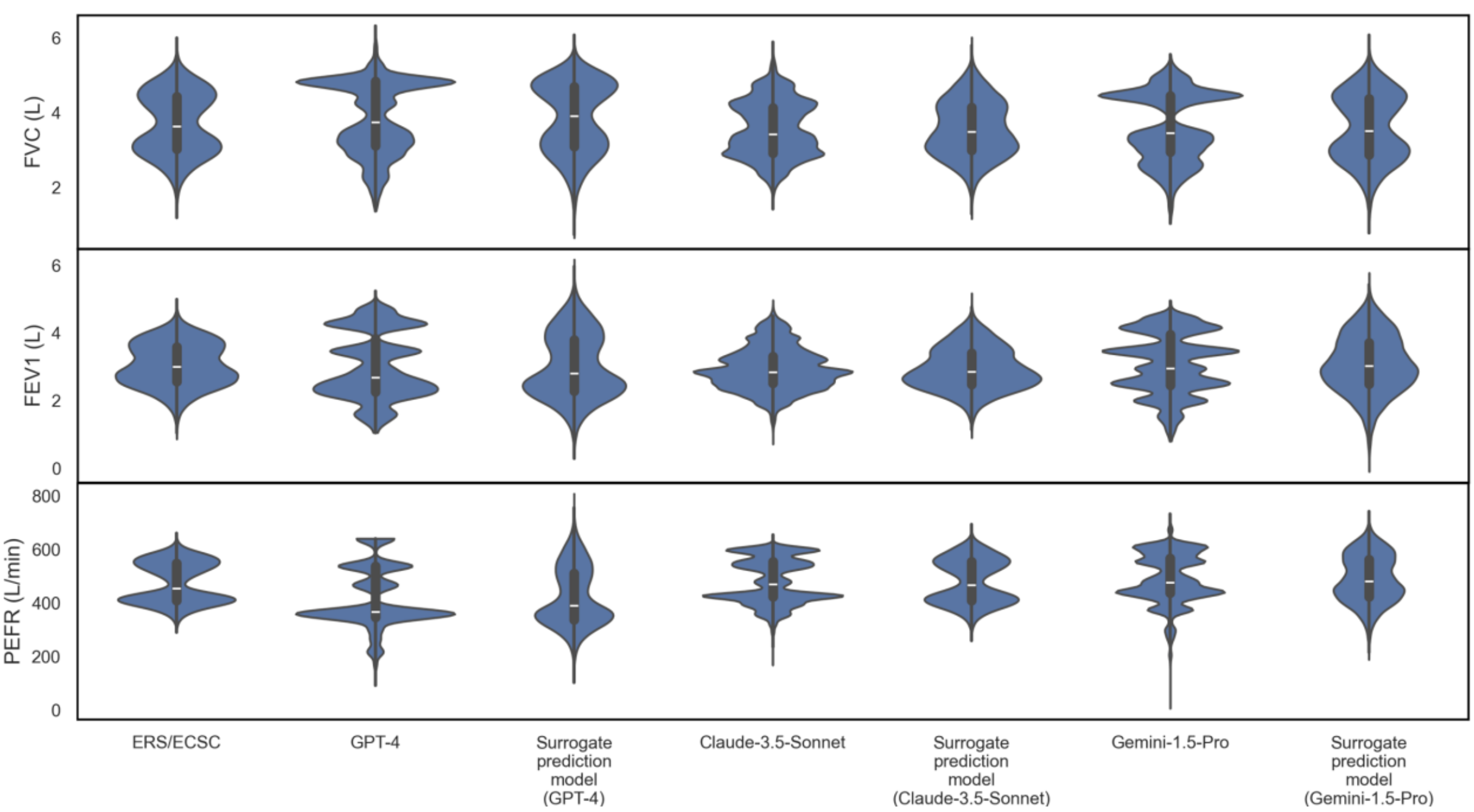

**Supplementary Fig. S7. Violin plots of FVC, FEV1, and PEFR derived from the ERS/ECSC reference equations[29], direct LLM predictions, and their corresponding surrogate models (Experiment 4).**

FVC: forced vital capacity, FEV1: forced expiratory volume in one second, PEFR: peak expiratory flow rate, ERS/ECSC: European Respiratory Society/European Coal and Steel Community, LLM: large language model

# Supplementary tables

**Supplementary Table S1. Linear regression coefficients (Experiment 1: Cardiovascular disease risk prediction).** All linear regression models were valid, with the independent variables explaining 68.3% to 87.6% of the variation in the dependent variable ($R^2$ = 0.683–0.876). All cases achieved overall statistical significance (F-test, P<0.001).

| Variables | Coefficient (95% CI) | | |
|---|---|---|---|
| | GPT-4 | Claude-3.5-Sonnet | Gemini-1.5-Pro |
| **Intercept** | -65.243 (-65.691 – -64.794) | -55.757 (-56.090 – -55.425) | -68.416 (-68.956 – -67.876) |
| **Age, years** | 0.784 (0.782 – 0.786) | 0.630 (0.629 – 0.632) | 0.622 (0.620 – 0.625) |
| **Sex** | 6.240 (6.186 – 6.294) | 3.061 (3.021 – 3.101) | 2.880 (2.815 – 2.945) |
| **Diabetes** | 14.562 (14.508 – 14.616) | 8.338 (8.297 – 8.378) | 4.899 (4.833 – 4.964) |
| **Anti-hypertensive treatment** | 9.031 (8.977 – 9.085) | 4.734 (4.694 – 4.774) | 4.685 (4.620 – 4.750) |
| **Dyslipidemia** | 7.294 (7.239 – 7.348) | 2.919 (2.879 – 2.959) | 3.161 (3.096 – 3.227) |
| **History of AF** | 7.029 (6.975 – 7.084) | 7.627 (7.586 – 7.667) | 6.958 (6.892 – 7.023) |
| **CKD** | 12.363 (12.308 – 12.417) | 4.955 (4.914 – 4.995) | 5.339 (5.273 – 5.404) |
| **FHCVD** | 2.922 (2.867 – 2.976) | 2.522 (2.482 – 2.563) | 3.193 (3.128 – 3.258) |
| **Smoking status** | | | |
| **Non-smoker** | Reference | Reference | Reference |
| **Ex-smoker** | 1.057 (0.991 – 1.124) | 0.935 (0.886 – 0.984) | 1.283 (1.203 – 1.363) |
| **Current smoker** | 5.418 (5.352 – 5.485) | 6.060 (6.011 – 6.109) | 2.560 (2.480 – 2.640) |
| **SBP, mmHg** | 0.084 (0.083 – 0.085) | 0.098 (0.097 – 0.099) | 0.123 (0.122 – 0.125) |
| **DBP, mmHg** | 0.017 (0.015 – 0.020) | 0.054 (0.052 – 0.055) | 0.047 (0.045 – 0.050) |
| **BMI, kg/m$^2$** | 0.093 (0.088 – 0.098) | 0.035 (0.031 – 0.039) | 0.134 (0.128 – 0.141) |
| **Waist–hip ratio** | 0.426 (0.137 – 0.714) | 3.948 (3.734 – 4.162) | -0.313 (-0.660 – 0.034) |
| **HDL, mg/dL** | -0.010 (-0.012 – -0.009) | -0.057 (-0.057 – -0.056) | -0.012 (-0.014 – -0.011) |
| **LDL, mg/dL** | 0.040 (0.039 – 0.041) | 0.023 (0.023 – 0.024) | 0.044 (0.043 – 0.045) |
| **Triglycerides, mg/dL** | 0.010 (0.010 – 0.010) | 0.004 (0.003 – 0.004) | 0.009 (0.008 – 0.009) |
| **HbA1c, %** | 0.506 (0.488 – 0.523) | 0.532 (0.519 – 0.545) | 0.615 (0.593 – 0.636) |
| **Creatinine, mg/dL** | 1.195 (1.117 – 1.273) | 0.940 (0.882 – 0.998) | 2.428 (2.334 – 2.522) |
| **Uric acid, mg/dL** | 0.035 (0.014 – 0.056) | 0.053 (0.038 – 0.069) | -0.046 (-0.071 – -0.021) |
| **CRP, mg/dL** | 0.258 (0.186 – 0.329) | 0.490 (0.437 – 0.543) | 0.661 (0.575 – 0.748) |

CI: confidence interval, HDL: high-density lipoprotein, LDL: low-density lipoprotein, SBP: systolic blood pressure, DBP: diastolic blood pressure, BMI: body mass index, AF: atrial fibrillation, CKD: chronic kidney disease, FHCVD: family history of cardiovascular disease in first-degree relatives, HbA1c: hemoglobin A1C, CRP: C-reactive protein.

**Supplementary Table S2. LLM prompt and response examples.**

| | **LLM prompt example** | **LLM response examples** |
|---|---|---|
| **Experiment 1: Cardiovascular disease risk prediction** | Estimate the risk (in percentages) of developing a cardiovascular disease within 10 years for the person below.<br><br>64 year old female, without diabetes, without hypertension, with dyslipidemia, without chronic kidney disease, without history of atrial fibrillation, with family history of cardiovascular disease in first degree relatives, non smoker, HDL 42.3 mg/dL, LDL 159.5 mg/dL, triglycerides 55.0 mg/dL, HbA1c 5.52 %, creatinine 0.84 mg/dL, urate 5.71 mg/dL, C-reactive protein 0.39 mg/dL, systolic blood pressure 122 mmHg, diastolic blood pressure 80 mmHg, BMI 23.8, waist to hip ratio 0.85<br><br>Please answer exactly in the format below, without blank lines, and no further information or answer is required.<br>Risk percentage=(in percentages, round to one decimal place) | Response 1:<br><br>Risk percentage=9.5%<br><br>Response 2:<br><br>Risk percentage=9.5%<br><br>Response 3:<br><br>Risk percentage=9.5% |
| **Experiment 2: Estimation of glomerular filtration rate** | What is the estimated glomerular filtration rate (eGFR) for the person below?<br><br>58 year old White female, serum creatinine 0.8 mg/dL<br><br>Please answer exactly in the format below, without blank lines, and no further information or answer is required.<br>eGFR=(in mL/min/1.73 $m^2$, round to two decimal places) | Response 1:<br><br>eGFR=81.53 mL/min/1.73 $m^2$<br><br>Response 2:<br><br>eGFR=85.35 mL/min/1.73 $m^2$<br><br>Response 3:<br><br>eGFR=81.27 mL/min/1.73 $m^2$ |
| **Experiment 3: Stroke risk prediction** | Estimate the risk (in percentages) of developing stroke within 10 years for the person below.<br><br>45 year old female, without diabetes, with hypertension, without dyslipidemia, with chronic kidney disease, without congestive heart failure, without coronary artery disease, without peripheral vascular disease, without history of atrial fibrillation, without history of previous stroke, with family history of cardiovascular disease in first degree relatives, current smoker, HDL 59.8 mg/dL, LDL 90.7 mg/dL, triglycerides 75.8 mg/dL, HbA1c 5.63 %, systolic blood pressure 168 mmHg, diastolic blood pressure 108 mmHg, BMI 22.73<br><br>Please answer exactly in the format below, without blank lines, and no further information or answer is required.<br>Risk percentage=(in percentages, round to one decimal place) | Response 1:<br><br>Risk percentage=14.1%<br><br>Response 2:<br><br>Risk percentage=8.2%<br><br>Response 3:<br><br>Risk percentage=5.2% |
| **Experiment 4: Estimation of pulmonary function** | Estimate Forced Vital Capacity (FVC) for the person below.<br><br>56 year old Black male, height 172.6 cm<br><br>Please answer exactly in the format below, without blank lines, and no further information or answer is required.<br>FVC=(in liters, round to two decimal places) | Response 1:<br><br>FVC=4.15 liters<br><br>Response 2:<br><br>FVC=4.10 liters<br><br>Response 3:<br><br>FVC=4.12 liters |

LLM: large language model, HDL: high-density liproptein, LDL: low-density lipoprotein, HbA1c: hemoglobin A1c, BMI: body mass index.

**Supplementary Table S3. Linear regression coefficients for various input variable combinations (Experiment 1, GPT-4).** "Surrogate prediction model (+None)" denotes a baseline using PCE variables for surrogate prediction model development. "+B" adds biochemical markers (LDL, triglycerides, HbA1c, creatinine, uric acid, CRP), "+P" adds physical measurements (DBP, BMI, waist-hip ratio), and "+H" adds medical history (dyslipidemia, AF, CKD, FHCVD). The independent variables explained between 81.6% and 88.8% of the variation in the dependent variable (R-squared ranging from 0.816 to 0.888) for the various input variable combinations. All input variable combinations achieved overall statistical significance (*P*<.001 in the F-test).

| | **Linear regression coefficients** | | | | | | | |
|---|---|---|---|---|---|---|---|---|
| **Variables** | **Surrogate prediction model (+B, P, H) (95% CI)** | **Surrogate prediction model (+H) (95% CI)** | **Surrogate prediction model (+P, H) (95% CI)** | **Surrogate prediction model (+B, H) (95% CI)** | **Surrogate prediction model (+B, P) (95% CI)** | **Surrogate prediction model (+B) (95% CI)** | **Surrogate prediction model (+P) (95% CI)** | **Surrogate prediction model (+None) (95% CI)** |
| **Intercept** | -65.243 (-65.691 – -64.794) | -63.636 (-63.903 – -63.369) | -67.249 (-67.661 – -66.838) | -62.770 (-63.081 – -62.459) | -63.612 (-64.100 – -63.124) | -63.445 (-63.793 – -63.097) | -67.425 (-67.874 – -66.976) | -65.234 (-65.522 – -64.947) |
| **Age, years** | 0.784 (0.782 – 0.786) | 0.806 (0.804 – 0.808) | 0.788 (0.786 – 0.790) | 0.808 (0.806 – 0.810) | 0.830 (0.829 – 0.832) | 0.864 (0.862 – 0.867) | 0.827 (0.824 – 0.829) | 0.854 (0.852 – 0.856) |
| **Sex (male)** | 6.240 (6.186 – 6.294) | 8.376 (8.319 – 8.433) | 6.626 (6.573 – 6.679) | 6.090 (6.033 – 6.147) | 8.640 (8.581 – 8.700) | 8.668 (8.604 – 8.732) | 11.617 (11.558 – 11.675) | 13.511 (13.448 – 13.574) |
| **Diabetes** | 14.562 (14.508 – 14.616) | 14.704 (14.637 – 14.761) | 15.904 (15.851 – 15.957) | 14.677 (14.621 – 14.734) | 12.532 (12.473 – 12.592) | 12.205 (12.141 – 12.269) | 11.633 (11.575 – 11.691) | 10.798 (10.735 – 10.861) |
| **Anti-hypertensive treatment** | 9.031 (8.977 – 9.085) | 8.210 (8.153 – 8.267) | 8.692 (8.639 – 8.746) | 9.075 (9.018 – 9.131) | 5.595 (5.535 – 5.654) | 5.533 (5.469 – 5.597) | 4.105 (4.046 – 4.163) | 3.791 (3.728 – 3.854) |
| **Dyslipidemia** | 7.294 (7.239 – 7.348) | 7.405 (7.348 – 7.462) | 7.891 (7.838 – 7.944) | 7.486 (7.429 – 7.542) | N/A | N/A | N/A | N/A |
| **History of AF** | 7.029 (6.975 – 7.084) | 6.772 (6.715 – 6.829) | 7.695 (7.642 – 7.748) | 6.960 (6.904 – 7.017) | N/A | N/A | N/A | N/A |
| **CKD** | 12.363 (12.308 – 12.417) | 11.141 (11.084 – 11.198) | 12.794 (12.741 – 12.847) | 12.069 (12.012 – 12.125) | N/A | N/A | N/A | N/A |
| **FHCVD** | 2.922 (2.867 – 2.976) | 2.300 (2.243 – 2.357) | 2.634 (2.581 – 2.687) | 2.779 (2.723 – 2.836) | N/A | N/A | N/A | N/A |
| **Smoking status** | | | | | | | | |
| **Non-smoker** | Reference | Reference | Reference | Reference | Reference | Reference | Reference | Reference |
| **Ex-smoker** | 1.057 (0.991 – 1.124) | 1.054 (0.984 – 1.123) | 1.082 (1.017 – 1.147) | 0.923 (0.854 – 0.993) | 2.820 (2.747 – 2.892) | 3.335 (3.256 – 3.414) | 2.789 (2.718 – 2.861) | 2.601 (2.524 – 2.678) |
| **Current smoker** | 5.418 (5.352 – 5.485) | 4.644 (4.574 – 4.714) | 5.052 (4.987 – 5.117) | 5.352 (5.282 – 5.421) | 9.278 (9.205 – 9.351) | 10.138 (10.059 – 10.216) | 8.829 (8.758 – 8.900) | 9.014 (8.937 – 9.091) |
| **SBP, mmHg** | 0.084 (0.083 – 0.085) | 0.125 (0.124 – 0.127) | 0.136 (0.135 – 0.138) | 0.087 (0.085 – 0.088) | 0.096 (0.095 – 0.098) | 0.101 (0.099 – 0.102) | 0.142 (0.141 – 0.143) | 0.139 (0.138 – 0.141) |
| **DBP, mmHg** | 0.017 (0.015 – 0.020) | N/A | 0.029 (0.027 – 0.031) | N/A | 0.015 (0.013 – 0.018) | N/A | 0.014 (0.012 – 0.016) | N/A |
| **BMI, kg/m$^2$** | 0.093 (0.088 – 0.098) | N/A | 0.125 (0.120 – 0.131) | N/A | 0.108 (0.103 – 0.114) | N/A | 0.171 (0.165 – 0.176) | N/A |
| **Waist–hip ratio** | 0.426 (0.137 – 0.714) | N/A | 0.126 (-0.156 – 0.407) | N/A | 0.327 (0.011 – 0.642) | N/A | 1.842 (1.532 – 2.152) | N/A |
| **Total cholesterol, mg/dL** | N/A | 0.034 (0.034 – 0.035) | 0.032 (0.031 – 0.032) | N/A | N/A | N/A | 0.042 (0.041 – 0.042) | 0.047 (0.047 – 0.048) |
| **HDL cholesterol, mg/dL** | -0.010 (-0.012 – -0.009) | -0.064 (-0.065 – -0.062) | -0.061 (-0.062 – -0.060) | -0.012 (-0.013 – -0.011) | -0.008 (-0.009 – -0.007) | -0.013 (-0.015 – -0.012) | -0.073 (-0.075 – -0.072) | -0.094 (-0.096 – -0.092) |
| **LDL cholesterol, mg/dL** | 0.040 (0.039 – 0.041) | N/A | N/A | 0.042 (0.041 – 0.043) | 0.056 (0.056 – 0.057) | 0.061 (0.061 – 0.062) | N/A | N/A |
| **Triglycerides, mg/dL** | 0.010 (0.010 – 0.010) | N/A | N/A | 0.011 (0.010 – 0.011) | 0.014 (0.014 – 0.014) | 0.015 (0.015 – 0.015) | N/A | N/A |
| **HbA1c, %** | 0.506 (0.488 – 0.523) | N/A | N/A | 0.421 (0.403 – 0.439) | 0.484 (0.465 – 0.504) | 0.441 (0.420 – 0.461) | N/A | N/A |
| **Creatinine, mg/dL** | 1.195 (1.117 – 1.273) | N/A | N/A | 1.118 (1.036 – 1.199) | 0.875 (0.789 – 0.961) | 1.067 (0.974 – 1.160) | N/A | N/A |
| **Uric acid, mg/dL** | 0.035 (0.014 – 0.056) | N/A | N/A | 0.049 (0.028 – 0.071) | 0.064 (0.041 – 0.087) | 0.047 (0.022 – 0.072) | N/A | N/A |
| **CRP, mg/dL** | 0.258 (0.186 – 0.329) | N/A | N/A | 0.250 (0.176 – 0.325) | 0.382 (0.304 – 0.461) | 0.288 (0.203 – 0.373) | N/A | N/A |

PCE: Pooled Cohort Equations, LDL: low-density lipoprotein, HbA1c: hemoglobin A1c, CRP: C-reactive protein, DBP: diastolic blood pressure, BMI: body mass index, AF: atrial fibrillation, CKD: chronic kidney disease, FHCVD: family history of cardiovascular disease in first-degree relatives, CI: confidence interval, HDL: high-density lipoprotein, SBP: systolic blood pressure, CRP: C-reactive protein.

**Supplementary Table S4. Baseline characteristics of the UKB dataset (Experiment 1).**

| | **Low risk by PCE (n = 215,844 [60.4%])** | **Moderate risk by PCE (n = 114,911 [32.2%])** | **High risk by PCE (n = 26,358 [7.4%])** | **Overall (n = 357,113)** | ***P*-value** |
|---|---|---|---|---|---|
| **10-year MACE incidence (%)** | 6,400 (2.97) | 11,542 (10.04) | 4,924 (18.68) | 22,866 (6.40) | <0.001 |
| **Age, years [IQR]** | 52 [46–58] | 62 [59–66] | 66 [64–68] | 57 [50–63] | <0.001 |
| **Sex (male) (%)** | 59,529 (27.58) | 75,819 (65.98) | 24,308 (92.22) | 159,656 (44.71) | <0.001 |
| **Diabetes (%)** | 3,684 (1.71) | 6,194 (5.39) | 6,272 (23.80) | 16,150 (4.52) | <0.001 |
| **Anti-hypertensive treatment (%)** | 19,632 (9.10) | 33,242 (28.93) | 12,908 (48.97) | 65,782 (18.42) | <0.001 |
| **Dyslipidemia (%)** | 16,266 (7.54) | 21,830 (19.00) | 7,047 (26.74) | 45,143 (12.64) | <0.001 |
| **History of AF (%)** | 1,350 (0.63) | 2,277 (1.98) | 716 (2.72) | 4,343 (1.22) | <0.001 |
| **CKD (%)** | 1,307 (0.61) | 1,714 (1.49) | 648 (2.46) | 3,669 (1.03) | <0.001 |
| **FHCVD (%)** | 88,316 (40.92) | 50,748 (44.16) | 10,936 (41.49) | 150,000 (42.00) | <0.001 |
| **Smoking status** | | | | | <0.001 |
| **Non-smoker (%)** | 134,375 (62.26) | 55,223 (48.06) | 9,227 (35.01) | 198,825 (55.68) | |
| **Ex-smoker (%)** | 67,259 (31.16) | 43,260 (37.65) | 11,048 (41.92) | 121,567 (34.04) | |
| **Current smoker (%)** | 14,210 (6.58) | 16,428 (14.30) | 6,083 (23.08) | 36,721 (10.28) | |
| **SBP, mmHg [IQR]** | 130.0 [120.0–141.0] | 144.0 [133.5–156.0] | 156.5 [146.0–169.0] | 136.0 [124.5–149.5] | <0.001 |
| **DBP, mmHg [IQR]** | 80.0 [73.5–87.0] | 84.5 [78.0–91.0] | 87.5 [81.0–94.5] | 82.0 [75.5–89.0] | <0.001 |
| **BMI, kg/m$^2$ [IQR]** | 26.0 [23.5–29.2] | 27.4 [25.0–30.3] | 28.2 [25.9–31.1] | 26.6 [24.1–29.8] | <0.001 |
| **Waist-hip ratio [IQR]** | 0.83 [0.78–0.90] | 0.91 [0.86–0.96] | 0.95 [0.91–0.99] | 0.87 [0.80–0.93] | <0.001 |
| **Total cholesterol, mg/dL [IQR]** | 219.6 [193.5–247.8] | 224.2 [194.1–254.9] | 214.0 [180.6–246.3] | 220.6 [192.9–249.9] | <0.001 |
| **HDL, mg/dL [IQR]** | 57.8 [48.7–68.3] | 51.3 [43.6–60.8] | 45.5 [39.2–53.2] | 54.6 [45.8–65.2] | <0.001 |
| **LDL, mg/dL [IQR]** | 135.6 [115.5–157.4] | 142.2 [118.9–165.7] | 136.6 [110.4–161.8] | 137.7 [116.1–160.5] | <0.001 |
| **Triglycerides, mg/dL [IQR]** | 116.0 [83.8–166.3] | 151.9 [108.7–214.9] | 176.4 [125.9–246.8] | 130.8 [92.2–189.3] | <0.001 |
| **HbA1c, % [IQR]** | 5.31 [5.10–5.53] | 5.44 [5.22–5.68] | 5.58 [5.32–6.01] | 5.36 [5.14–5.60] | <0.001 |
| **Creatinine, mg/dL [IQR]** | 0.75 [0.67–0.86] | 0.85 [0.74–0.96] | 0.90 [0.80–1.01] | 0.79 [0.69–0.91] | <0.001 |
| **Uric acid, mg/dL [IQR]** | 4.69 [3.9–5.6] | 5.55 [4.7–6.4] | 5.93 [5.1–6.8] | 5.06 [4.2–6.0] | <0.001 |
| **CRP, mg/dL [IQR]** | 0.12 [0.06–0.25] | 0.15 [0.08–0.30] | 0.19 [0.10–0.35] | 0.13 [0.07–0.27] | <0.001 |
| **Surrogate prediction model score [IQR]** | 6.7 [0.7–11.8] | 20.6 [16.7–26.6] | 32.5 [26.2–40.7] | 13.0 [4.9–21.3] | <0.001 |
| **PCE score [IQR]** | 3.0 [1.5–4.9] | 11.8 [9.4–15.0] | 24.2 [21.8–28.3] | 5.6 [2.4–11.3] | <0.001 |
| **FRS [IQR]** | 7.3 [4.6–10.4] | 19.8 [16.2–24.3] | 37.8 [32.0–44.6] | 11.4 [6.4–19.2] | <0.001 |

UKB: United Kingdom Biobank, PCE: Pooled Cohort Equations, MACE: major adverse cardiovascular event, IQR: interquartile range, AF: atrial fibrillation, CKD: chronic kidney disease, FHCVD: family history of cardiovascular disease in first-degree relatives, SBP: systolic blood pressure, DBP: diastolic blood pressure, BMI: body mass index, HDL: high-density lipoprotein, LDL: low-density lipoprotein, HbA1c: hemoglobin A1c, CRP: C-reactive protein, FRS: Framingham Risk Score.

**Supplementary Table S5. Predicted 10-year cardiovascular disease risk estimates from FRS, PCE, and surrogate models (Experiment 1).**

| | **FRS** | **PCE** | **Surrogate model for GPT-4** | **Surrogate model for Claude-3.5-Sonnet** | **Surrogate model for Gemini-1.5-Pro** |
|---|---|---|---|---|---|
| **Predicted risk, %, mean ± std** | 14.2 ± 10.7 | 7.9 ± 7.3 | 13.7 ± 12.5 | 11.8 ± 9.1 | 7.9 ± 9.2 |

FRS: Framingham Risk Score, PCE: Pooled Cohort Equations, std: standard deviation, IQR: interquartile range

**Supplementary Table S6. Performance of the surrogate prediction model across various input variable combinations (Experiment 1, GPT-4, C-index).** The C-index of PCE is 0.727 (95% CI; 0.724–0.730) and that of the FRS was 0.720 (95% CI; 0.718–0.724). "Surrogate prediction model (+None)" denotes a baseline using PCE variables for surrogate prediction model development. "+B" adds biochemical markers (LDL, triglycerides, HbA1c, creatinine, uric acid, CRP), "+P" adds physical measurements (DBP, BMI, waist-hip ratio), and "+H" adds medical history (dyslipidemia, AF, CKD, FHCVD).

| Input combinations | C-index (95% CI) | Differences in C-index with the PCE (95% CI) | Differences in C-index with the FRS (95% CI) |
|---|---|---|---|
| **Surrogate prediction model (+B, P, H)** | 0.736 (0.734 – 0.739) | 0.010 (0.008 – 0.011) | 0.016 (0.014 – 0.018) |
| **Surrogate prediction model (+H)** | 0.736 (0.733 – 0.739) | 0.009 (0.007 – 0.011) | 0.015 (0.014 – 0.017) |
| **Surrogate prediction model (+P, H)** | 0.735 (0.732 – 0.738) | 0.008 (0.007 – 0.010) | 0.015 (0.013 – 0.017) |
| **Surrogate prediction model (+B, H)** | 0.735 (0.732 – 0.738) | 0.008 (0.007 – 0.010) | 0.015 (0.013 – 0.017) |
| **Surrogate prediction model (+B, P)** | 0.732 (0.729 – 0.735) | 0.005 (0.004 – 0.006) | 0.011 (0.010 – 0.013) |
| **Surrogate prediction model (+B)** | 0.730 (0.727 – 0.733) | 0.003 (0.002 – 0.004) | 0.010 (0.009 – 0.011) |
| **Surrogate prediction model (+P)** | 0.729 (0.726 – 0.732) | 0.002 (0.001 – 0.003) | 0.008 (0.007 – 0.009) |
| **Surrogate prediction model (+None)** | 0.724 (0.721 – 0.727) | -0.003 (-0.004 – -0.002) | 0.004 (0.003 – 0.005) |

PCE: Pooled Cohort Equations, CI: confidence interval, FRS: Framingham Risk Score, LDL: low-density lipoprotein, HbA1c: hemoglobin A1c, CRP: C-reactive protein, DBP: diastolic blood pressure, BMI: body mass index, AF: atrial fibrillation, CKD: chronic kidney disease, FHCVD: family history of cardiovascular disease in first-degree relatives.

**Supplementary Table S7. Comparison of inference time between the surrogate model and API-based LLM prompting (Experiment 1, 1000 random samples).** P-values indicate comparisons against the inference time of the surrogate model. Inference time was measured on a system equipped with an Intel i9-13900K CPU.

| | Surrogate model | GPT-4 | Claude-3.5-Sonnet | Gemini-1.5-Pro |
|---|---|---|---|---|
| **Time, microseconds, median [IQR]** | 841 [750–850] | 1,024,263 [853,573–1,259,588], P<0.001 | 1,823,891 [1,419,691–2,390,993], P<0.001 | 774,390 [720,906–899,867], P<0.001 |

API: application programming interface, LLM: large language model, CPU: central processing unit

**Supplementary Table S8. Linear regression coefficients (Experiment 2: Estimation of glomerular filtration rate).** All linear regression models were valid, with the independent variables explaining 84.0% to 95.7% of the variation in the dependent variable ($R^2$ = 0.840–0.957). All cases achieved overall statistical significance (F-test, P<0.001).

| Independent variable | GPT-4 | Claude-3.5-Sonnet | Gemini-1.5-Pro | 2021 CKD-EPI eGFRcr equation | 2009 CKD-EPI eGFRcr equation |
|---|---|---|---|---|---|
| **Intercept $\mu$ (95% CI)** | 156 (154 – 157) | 137 (136 – 137) | 178 (177 – 179) | 142 (139 – 144) | 141 (139 – 144) |
| **Coefficient $\beta_1$ for Scr (95% CI)** | F: -0.380 (-0.396 – -0.363) M: -0.390 (-0.401 – -0.380) | F: -0.856 (-0.865 – -0.846) M: -0.677 (-0.684 – -0.671) | F: -0.274 (-0.285 – -0.263) M: -0.495 (-0.502 – -0.488) | F: -0.241 (-0.344 – -0.138) M: -0.302 (-0.403 – -0.202) | F: -0.329 (-0.428 – -0.230) M: -0.411 (-0.508 – -0.314) |
| **Coefficient $\beta_2$ for Scr (95% CI)** | -1.054 (-1.063 – -1.044) | -1.085 (-1.090 – -1.079) | -1.355 (-1.362 – -1.349) | -1.200 (-1.211 – -1.189) | -1.209 (-1.220 – -1.198) |
| **Coefficient $\gamma$ for age (95% CI)** | 0.9919 (0.9917 – 0.9920) | 0.9934 (0.9933 – 0.9935) | 0.9894 (0.9893 – 0.9895) | 0.9938 (0.9935 – 0.9942) | 0.9929 (0.9925 – 0.9933) |
| **Coefficient $\delta$ for female (95% CI)** | 1.187 (1.181 – 1.193) | 1.054 (1.050 – 1.057) | 1.056 (1.052 – 1.059) | 1.012 (1.000 – 1.023) | 1.018 (1.007 – 1.029) |
| **Coefficient $\varepsilon$ for black race (95% CI)** | 1.153 (1.149 – 1.158) | 1.227 (1.224 – 1.230) | 1.008 (1.005 – 1.010) | None | 1.159 (1.144 – 1.170) |

CKD-EPI: Chronic Kidney Disease Epidemiology Collaboration, eGFRcr: glomerular filtration rate estimated with the use of creatinine, CI: confidence interval, F: female, M: male, Scr: serum creatinine

**Supplementary Table S9. Estimated GFR values from CKD-EPI equations, direct LLM predictions, and corresponding surrogate models within the simulated dataset (Experiment 2).**

| | 2009 CKD-EPI eGFRcr formula | 2021 CKD-EPI eGFRcr formula | GPT-4 | Surrogate model for GPT-4 | Claude-3.5-Sonnet | Surrogate model for Claude-3.5-Sonnet | Gemini-1.5-Pro | Surrogate model for Gemini-1.5-Pro |
|---|---|---|---|---|---|---|---|---|
| **Estimated GFR, mL/min/1.73m$^2$, mean ± std** | 92.67 ± 26.25 | 97.43 ± 30.43 | 111.65 ± 31.94 | 110.99 ± 31.84 | 108.85 ± 36.78 | 108.81 ± 37.65 | 96.23 ± 33.03 | 96.02 ± 33.09 |

GFR: glomerular filtration rate, CKD-EPI: Chronic Kidney Disease Epidemiology Collaboration, LLM: large language model, eGFRcr: glomerular filtration rate estimated with the use of creatinine, std: standard deviation

**Supplementary Table S10. Linear regression coefficients (Experiment 3: Stroke risk prediction).** All linear regression models were valid, with the independent variables explaining 80.1% to 87.6% of the variation in the dependent variable ($R^2$ = 0.801–0.876). All cases achieved overall statistical significance (F-test, $P<0.001$).

| Variables | Coefficient (95% CI) | | |
|---|---|---|---|
| | **GPT-4** | **Claude-3.5-Sonnet** | **Gemini-1.5-Pro** |
| **Intercept** | -39.164 (-39.466 – -38.861) | -47.401 (-47.634 – -47.169) | -38.015 (-38.194 - -37.836) |
| **Age, years** | 0.454 (0.452 – 0.456) | 0.375 (0.374 – 0.376) | 0.354 (0.353 – 0.355) |
| **Male sex** | 1.641 (1.601 – 1.681) | 0.710 (0.679 – 0.740) | 1.233 (1.210 – 1.257) |
| **Diabetes** | 7.286 (7.247 – 7.326) | 4.973 (4.943 – 5.004) | 2.470 (2.446 – 2.494) |
| **Anti-hypertensive treatment** | 2.891 (2.851 – 2.931) | 4.377 (4.346 – 4.407) | 3.646 (3.622 – 3.669) |
| **Dyslipidemia** | 3.636 (3.596 – 3.676) | 0.839 (0.808 – 0.870) | 1.038 (1.014 – 1.061) |
| **History of AF** | 3.966 (3.926 – 4.006) | 9.826 (9.795 – 9.856) | 6.147 (6.124 – 6.171) |
| **CKD** | 5.504 (5.464 – 5.544) | 5.468 (5.437 – 5.498) | 3.924 (3.900 – 3.948) |
| **CHF** | 8.126 (8.086 – 8.166) | 5.306 (5.276 – 5.337) | 4.009 (3.985 – 4.032) |
| **CAD** | 5.219 (5.179 – 5.258) | 3.953 (3.923 – 3.984) | 3.870 (3.847 – 3.894) |
| **PVD** | 3.212 (3.172 – 3.252) | 4.794 (4.763 – 4.824) | 2.870 (2.846 – 2.893) |
| **FHCVD** | 0.987 (0.948 – 1.027) | 0.860 (0.829 – 0.891) | 0.610 (0.586 – 0.633) |
| **Smoking status** | | | |
| **Non-smoker** | Reference | Reference | Reference |
| **Ex-smoker** | 0.216 (0.168 – 0.265) | 1.165 (1.127 – 1.202) | 0.574 (0.545 – 0.603) |
| **Current smoker** | 1.974 (1.924 – 2.023) | 3.796 (3.758 – 3.833) | 1.122 (1.094 – 1.152) |
| **SBP, mmHg** | 0.068 (0.067 – 0.069) | 0.120 (0.120 – 0.121) | 0.082 (0.081 – 0.082) |
| **DBP, mmHg** | 0.012 (0.010 – 0.014) | 0.067 (0.065 – 0.068) | 0.048 (0.046 – 0.049) |
| **BMI, kg/m$^2$** | 0.038 (0.034 – 0.042) | 0.032 (0.029 – 0.035) | 0.031 (0.029 – 0.033) |
| **HDL, mg/dL** | -0.009 (-0.009 – -0.008) | -0.034 (-0.035 – -0.033) | -0.019 (-0.020 – 0.019) |
| **LDL, mg/dL** | 0.022 (0.021 – 0.023) | 0.010 (0.010 – 0.011) | 0.017 (0.017 – 0.017) |
| **Triglycerides, mg/dL** | 0.003 (0.003 – 0.003) | 0.002 (0.002 – 0.003) | 0.004 (0.004 – 0.004) |
| **HbA1c, %** | 0.180 (0.160 – 0.200) | 0.602 (0.587 – 0.618) | 0.190 (0.177 – 0.202) |

CI: confidence interval, AF: atrial fibrillation, CKD: chronic kidney disease, CHF: congestive heart failure, CAD: coronary artery disease, PVD: peripheral vascular disease, FHCVD: family history of cardiovascular disease in first-degree relatives, SBP: systolic blood pressure, DBP: diastolic blood pressure, BMI: body mass index, HDL: high-density lipoprotein, LDL: low-density lipoprotein, HbA1c: hemoglobin A1c.

**Supplementary Table S11. Predicted 10-year stroke risk estimates from the FSRP and surrogate models (Experiment 3).**

| | FSRP | Surrogate model for GPT-4 | Surrogate model for Claude-3.5-Sonnet | Surrogate model for Gemini-1.5-Pro |
|---|---|---|---|---|
| **Predicted risk, %, mean ± std** | 2.5 ± 3.3 | 5.1 ± 6.7 | 3.1 ± 6.9 | 9.9 ± 2.2 |

FSRP: Framingham Stroke Risk Profile, std: standard deviation

**Supplementary Table S12. Linear regression coefficients (Experiment 4: Estimation of pulmonary function).** All linear regression models were valid, with the independent variables explaining 72.1% to 97.9% of the variation in the dependent variable ($R^2$ = 0.721–0.979). All cases achieved overall statistical significance (F-test, $P<0.001$).

| Dependent variable | Independent variable | Coefficient (95% CI) | | |
|---|---|---|---|---|
| | | GPT-4 | Claude-3.5-Sonnet | Gemini-1.5-Pro |
| FVC, L (men) | Intercept | -2.481 (-2.653 – -2.309) | -5.587 (-5.650 – -5.523) | -3.544 (-3.717 – -3.372) |
| | Age, years | -0.024 (-0.025 – -0.024) | -0.025 (-0.025 – -0.025) | -0.029 (-0.029 – -0.028) |
| | Height, cm | 0.048 (0.047 – 0.049) | 0.064 (0.064 – 0.064) | 0.054 (0.053 – 0.055) |
| | Black race | -0.013 (-0.023 – -0.003) | -0.438 (-0.442 – -0.434) | -0.236 (-0.245 – -0.226) |
| FVC, L (women) | Intercept | -3.704 (-3.805 – -3.604) | -2.972 (-3.020 – -2.924) | -1.648 (-1.730 – -1.565) |
| | Age, years | -0.035 (-0.035 – -0.034) | -0.023 (-0.023 – -0.022) | -0.031 (-0.031 – -0.031) |
| | Height, cm | 0.053 (0.052 – 0.054) | 0.045 (0.044 – 0.045) | 0.038 (0.037 – 0.038) |
| | Black race | -0.162 (-0.168 – -0.156) | -0.357 (-0.360 – -0.354) | -0.161 (-0.166 – -0.156) |
| FEV1, L (men) | Intercept | -5.312 (-5.502 – -5.122) | -3.409 (-3.479 – -3.340) | -1.358 (-1.548 – -1.167) |
| | Age, years | -0.048 (-0.048 – -0.047) | -0.031 (-0.031 – -0.031) | -0.048 (-0.048 – -0.047) |
| | Height, cm | 0.066 (0.065 – 0.067) | 0.049 (0.049 – 0.049) | 0.044 (0.043 – 0.045) |
| | Black race | -0.263 (-0.274 – -0.253) | -0.439 (-0.443 – -0.435) | -0.359 (-0.370 – -0.349) |
| FEV1, L (women) | Intercept | -1.991 (-2.074 – -1.908) | -1.714 (-1.748 – -1.680) | 0.010 (-0.084 – 0.104) |
| | Age, years | -0.031 (-0.036 – -0.031) | -0.022 (-0.022 – -0.022) | -0.042 (-0.042 – -0.042) |
| | Height, cm | 0.036 (0.036 – 0.037) | 0.034 (0.033 – 0.034) | 0.029 (0.028 – 0.030) |
| | Black race | -0.074 (-0.079 – -0.069) | -0.285 (-0.287 – -0.283) | -0.162 (-0.167 – -0.156) |
| PEFR, L/min (men) | Intercept | -757.5 (-783.8 – -731.2) | -37.9 (-49.8 – -26.0) | 12.3 (-10.3 – 34.8) |
| | Age, years | -5.6 (-5.7 – -5.6) | -2.1 (-2.2 – -2.1) | -4.3 (-4.3 – -4.2) |
| | Height, cm | 8.9 (8.8 – 9.1) | 4.1 (4.0 – 4.2) | 4.4 (4.3 – 4.5) |
| | Black race | -27.8 (-29.3 – -26.3) | -38.5 (-39.2 – -37.8) | -6.7 (-7.9 – -5.4) |
| PEFR, L/min (women) | Intercept | 155.6 (139.3 – 172.0) | -15.2 (-23.5 – -7.0) | -21.6 (-36.7 – -6.6) |
| | Age, years | -3.2 (-3.2 – -3.2) | -2.0 (-2.0 – -2.0) | -3.6 (-3.6 – -3.5) |
| | Height, cm | 2.2 (2.1 – 2.3) | 3.3 (3.2 – 3.3) | 3.9 (3.8 – 4.0) |
| | Black race | -18.3 (-19.3 – -17.3) | -20.2 (-20.7 – -20.0) | -21.3 (-22.2 – -20.4) |

CI: confidence interval, FVC: forced vital capacity, FEV1: forced expiratory volume in one second, PEFR: peak expiratory flow rate.

**Supplementary Table S13. Estimated pulmonary function values from the ERS/ECSC reference equations[29], direct LLM predictions, and their corresponding surrogate models (Experiment 4).**

| | ERS/ECSC | GPT-4 | Surrogate model for GPT-4 | Claude-3.5-Sonnet | Surrogate model for Claude-3.5-Sonnet | Gemini-1.5-Pro | Surrogate model for Gemini-1.5-Pro |
|---|---|---|---|---|---|---|---|
| Estimated FVC, L, mean ± std | 3.70 ± 0.82 | 3.85 ± 0.96 | 3.85 ± 0.93 | 3.56 ± 0.74 | 3.56 ± 0.74 | 3.59 ± 0.91 | 3.59 ± 0.89 |
| Estimated FEV1, L, mean ± std | 3.05 ± 0.64 | 3.01 ± 0.97 | 3.01 ± 0.94 | 2.94 ± 0.63 | 2.94 ± 0.62 | 3.08 ± 0.88 | 3.08 ± 0.85 |
| Estimated PEFR, L/min, mean ± std | 474.9 ± 75.8 | 420.7 ± 112.8 | 420.7 ± 108.2 | 480.0 ± 83.0 | 480.0 ± 81.7 | 488.6 ± 92.0 | 488.6 ± 87.6 |

ERS/ECSC: European Respiratory Society/European Coal and Steel Community, LLM: large language model, FVC: forced vital capacity, FEV1: forced expiratory volume in one second, PEFR: peak expiratory flow rate, std: standard deviation

**Supplementary Table S14. Results of the temperature experiments.** Prior to conducting Experiment 1, a temperature experiment was performed using GPT-4. The lowest temperature setting yielded the highest response validity, predictive performance, and the smallest divergence from the actual 10-year MACE incidence, albeit with minimal differences. This was therefore selected as the optimal temperature.

| | **Temperatures** | | | | | |
|---|---|---|---|---|---|---|
| | **0.0** | **0.2** | **0.4** | **0.6** | **0.8** | **1.0** |
| **Response validity, %** | 99.843 | 99.838 | 99.730 | 99.455 | 99.239 | 99.100 |
| **Predictive performance of the surrogate prediction model for MACE, C-index** | 0.73632 | 0.73621 | 0.73626 | 0.73627 | 0.73614 | 0.73620 |
| **Average score of the surrogate prediction model, %** | 13.69 | 13.80 | 14.05 | 14.42 | 14.67 | 15.13 |

MACE: major adverse cardiovascular event.

**Supplementary Table S15. Variables extracted from the UKB.**

| Description | Category | Coding |
|---|---|---|
| Age at recruitment | Baseline characteristics | Field: 21022 |
| Sex | Baseline characteristics | Field: 31 |
| Diabetes diagnosed by doctor | Medical conditions | Field: 2443 |
| Medication for cholesterol, blood pressure or diabetes | Medication | Field: 6177 |
| Medication for cholesterol, blood pressure or diabetes, or exogenous hormones | Medication | Field: 6153 |
| Smoking status | Smoking | Field: 20116 |
| Systolic blood pressure | Blood pressure | Field: 4080 |
| Diastolic blood pressure | Blood pressure | Field: 4079 |
| Cholesterol | Blood chemistry | Field: 30690 |
| HDL cholesterol | Blood chemistry | Field: 30760 |
| LDL direct | Blood chemistry | Field: 30780 |
| Triglycerides | Blood chemistry | Field: 30870 |
| HbA1c | Blood chemistry | Field: 30750 |
| Creatinine | Blood chemistry | Field: 30700 |
| Urate | Blood chemistry | Field: 30880 |
| C-reactive protein | Blood chemistry | Field: 30710 |
| Ethnic background | Ethnicity | Field: 21000 |
| Illness of father | Family history | Field: 20107 |
| Illness of mother | Family history | Field: 20110 |
| Illness of siblings | Family history | Field: 20111 |
| Standing height | Body size measures | Field: 50 |
| Weight | Body size measures | Field: 21002 |
| Waist circumference | Body size measures | Field: 48 |
| Hip circumference | Body size measures | Field: 49 |
| Dyslipidemia, ICD-10: E78 | First occurrence | Field: 130814 |
| Acute myocardial infarction, ICD-10: I21 | First occurrence | Field: 131298 |
| Subsequent myocardial infarction, ICD-10: I22 | First occurrence | Field: 131300 |
| Certain current complications following acute myocardial infarction, ICD-10: I23 | First occurrence | Field: 131302 |
| Other acute ischemic heart disease, ICD-10: I24 | First occurrence | Field: 131304 |
| Chronic ischemic heart disease, ICD-10: I25 | First occurrence | Field: 131306 |
| Atrial fibrillation, ICD-10: I48 | First occurrence | Field: 131350 |
| Heart failure, ICD-10: I50 | First occurrence | Field: 131354 |
| Subarachnoid hemorrhage, ICD-10: I60 | First occurrence | Field: 131360 |
| Intracerebral hemorrhage, ICD-10: I61 | First occurrence | Field: 131362 |
| Other nontraumatic intracranial hemorrhage, ICD-10: I62 | First occurrence | Field: 131364 |
| Cerebral infarction, ICD-10: I63 | First occurrence | Field: 131366 |
| Stroke, not specified as hemorrhage or infarction, ICD-10: I64 | First occurrence | Field: 131368 |
| Atherosclerosis, ICD-10: I70 | First occurrence | Field: 131380 |
| Chronic kidney disease, ICD-10: N18 | First occurrence | Field: 132032 |

UKB: United Kingdom Biobank, HDL: high-density lipoprotein, LDL: low-density lipoprotein, HbA1c: hemoglobin A1C, ICD-10: International Classification of Diseases, 10th revision.